\documentclass[twocolumn,10pt,twoside,final]{asme2ej}
\usepackage{amsfonts}
\usepackage{graphicx}
\usepackage{amsfonts}
\usepackage{graphicx}
\usepackage{amsmath, amssymb, bm}
\usepackage[fixamsmath,disallowspaces]{mathtools}
\usepackage{xcolor}
\usepackage{fixmath}

\newtheorem{remark}{Remark}

\begin{document}

\parindent0pt \parskip2pt \setcounter{topnumber}{9} %
\setcounter{bottomnumber}{9} \renewcommand{\textfraction}{0.00001}

\renewcommand {\floatpagefraction}{0.999} \renewcommand{\textfraction}{0.01} %
\renewcommand{\topfraction}{0.999} \renewcommand{\bottomfraction}{0.99} %
\renewcommand{\floatpagefraction}{0.99} \setcounter{totalnumber}{9}

\title{Singularity-free Lie Group Integration and Geometrically Consistent Evaluation of Multibody System Models described in terms of Standard Absolute Coordinates
\vspace{-8mm}
} 
\author{
	{Andreas M\"uller}
	\affiliation{Johannes Kepler University, Linz, Austria\\
	a.mueller@jku.at}
} 

\maketitle 

\begin{abstract}
\textit{Abstract--} A classical approach to the MBS modeling is to use
absolute coordinates, i.e. a set of (possibly redundant) coordinates that
describe the absolute position and orientation of the individual bodies
w.r.t. to an inertial frame (IFR). A well-known problem for the time
integration of the equations of motion (EOM) is the lack of a
singularity-free parameterization of spatial motions, which is usually
tackled by using unit quaternions. 
\emph{Lie group integration methods} were proposed as alternative approach
to the singularity-free time integration. At the same time, \emph{Lie group
formulations of EOM} naturally respect the geometry of spatial motions
during integration. 
Lie group integration methods, operating directly on the configuration space
Lie group, are incompatible with standard formulations of the EOM, and
cannot be implemented in existing MBS simulation codes without a major
restructuring. 
The contribution of this paper is two-fold: 1) A framework for interfacing
Lie group integrators to standard EOM formulations is presented is
presented. It allows describing MBS in terms of various absolute coordinates
and at the same using Lie group integration schemes. 2) A method for
consistently incorporating the geometry of rigid body motions into the
evaluation of EOM in absolute coordinates integrated with standard vector
space integration schemes. 
The direct product group $SO\left( 3\right) \times {\mathbb{R}}^{3}$ and the
semidirect product group $SE\left( 3\right) $ are use for representing rigid
body motions. The key element is the \emph{local-global transitions (LGT)
transition map}, which facilitates the update of (global) absolute
coordinates in terms of the (local) coordinates on the Lie group. This LGT
map is specific to the absolute coordinates, the local coordinates on the
Lie group, and the Lie group used to represent rigid body configurations.

\end{abstract} 

\section{Introduction}

Complex multibody systems (MBS) are frequently modeled in terms of 'absolute
coordinates' describing the absolute position and orientation of the
individual bodies relative to a global inertial frame (IFR) \cite%
{ShabanaBook}. The corresponding equations of motions (EOM) form system of a
differential-algebraic equations (DAE), for which dedicated numerical
integration schemes exist. In the last decade, Lie group integration methods
for MBS have been developed and successfully applied to complex rigid body
systems where spatial motions are represented in a Lie group, but also to
systems with flexible bodies \cite%
{ArnoldBrulsCardona2015,ArnoldHante2017,BrulsCardona2010,BrulsCardonaArnold2012,CelledoniOwren2003,TerzeZlatarMueller2012,TerzeMueller2016,TerzeZlatarMueller2015}%
. Despite their clear benefits, Lie group methods for the time integration
of dynamic EOM have not found a wider application in MBS\ dynamics. The
apparent bottleneck of Lie group integration methods is that they are not
compatible with the standard formulation of MBS models in terms of absolute
coordinates. Typical choices for absolute coordinates are 3 positions and 3
angles, 3 positions and the 3 components of the rotation axis time angle
(the scaled rotation axis), or 3 positions and the 4 (dependent) components
of a unit quaternion. The latter is the standard approach to mitigate the
singularity problem encountered with any 3-parametric description of spatial
rotations at the expense of introducing additional constraints that must be
stabilized during time integration. Lie group methods, on the other hand,
use local coordinates to describe the motion of an MBS relative to its
configuration at the previous time step. Moreover, the main premise of the
Lie group methods is that no full turn rotation occurs during an integration
time step. The local parameters are specific to the Lie group used to
describe the rigid body motions. Rigid body motions form the Lie group $%
SE\left( 3\right) $, which should hence be used \cite%
{MMTCSpace2014,BIT2016,BIT2016Erratum,SonnevilleCardonaBruls2014,BorriTrainelliBottasso2000}%
. Yet the direct product group $SO\left( 3\right) \times {\mathbb{R}}^{3}$
is often used that does not represent rigid body motions, which is obvious
noting that rotations and translations are treated decoupled. Canonical
coordinates on $SE\left( 3\right) =SO\left( 3\right) \ltimes {\mathbb{R}}^{3}
$ are screw coordinates, while on $SO\left( 3\right) \times {\mathbb{R}}^{3}$
these are rotation/angle plus displacement vector. Non-canonical coordinates
on either Lie group are Gibbs-Rodrigues parameters, for instance giving rise
to algebraic parameterizations.

Lie group methods require a different formulation of MBS models and involve
a transition step that yields the MBS configuration due to the relative
motion within the time step, which will be referred to as local-global
transition step.

In this paper, an interfacing technique is proposed that allows describing
MBS in terms of standard absolute coordinates and at the same time using the
Lie group integration methods, where the direct as well as the semidirect
product representation of rigid body motions can be used. It admits
singularity-free integration using non-redundant absolute coordinates
without resorting to use dependent unit quaternions. This topic has not been
addressed in the literature with the exception of \cite%
{HolzingerGerstmayr2021}, which reported a Lie group method for the
integration of spatial rotations described in terms of global rotation
parameters. This paper provides the conceptual framework; the computational
aspects, and analysis of the computational efficiency of the various
combinations of absolute and local coordinates are subject of further
research. Arguably, the combination of unit quaternions and position vectors
as absolute coordinates with screw coordinates as local coordinates
(combination 1.a) in Sec. \ref{secLGT}), and with scaled rotation axis and
position vector (combination 1.c) in Sec. \ref{secLGT}) are the most
relevant. With the embedded Lie group integration, the geometry of the
underlying Lie group is inherited although the MBS configurations are
described by standard absolute coordinates. The geometry is encapsulated in
the LGT map, and regardless of the used absolute coordinates, the update of
rigid body configurations complies with this model. In particular, the
geometry of proper rigid body motions is respected when $SE\left( 3\right) $
is used. The paper is organized as follows. In section \ref{secVecDAE}, the
standard DAE formulation of the EOM for rigid body MBS are recalled. The
corresponding Lie group formulation is presented in section \ref{secEOMLie},
where the MBS dynamics is described on a general Lie group. Section \ref%
{secLieInt} summarizes the main aspects of Lie group integration methods
with emphasize on model evaluation within the integration step by means of
local coordinates. The two Lie groups $SO\left( 3\right) \times {\mathbb{R}}%
^{3}$ and $SE\left( 3\right) $, relevant for MBS modeling, and the most
relevant coordinates on these groups are recalled in section \ref%
{secCoordMaps}. The LGT map is introduced in section \ref{secEmbed}. In this
section, the interfacing of absolute coordinate models with Lie group
integrators, and the geometrically consistent integration of absolute
coordinate models with standard vector space integration schemes addressed.
The LGT map is derived explicitly for the most relevant combinations of
absolute and local coordinates. A short conclusion in section \ref%
{secConclusion} is followed by an appendix, where the BCH formula on $%
SO\left( 3\right) $ is derived, for completeness, and the LGT map relating
Rodrigues parameter and unit quaternions is presented.

\section{Equations of Motion in\ Vector Space DAE Form%
\label{secVecDAE}%
}

The EOM of a multibody system (comprising $N$ rigid bodies) in absolute
coordinates can be expressed in descriptor form as an index 2 DAE system%
\begin{eqnarray}
\mathbf{M}\left( \mathbf{q}\right) \dot{\mathbf{V}}+\mathbf{A}^{T}%
\bm{\lambda}
&=&\mathbf{Q}\left( \mathbf{q},\mathbf{V},t\right)   \label{EOMDyn} \\
\mathbf{V} &=&\mathbf{H}\left( \mathbf{q}\right) \dot{\mathbf{q}}
\label{EOMkin} \\
\mathbf{A}\left( \mathbf{q}\right) \mathbf{V} &=&\mathbf{0}  \label{VelConst}
\\
g\left( \mathbf{q}\right)  &=&\mathbf{0}  \label{GeomConst}
\end{eqnarray}%
where $\mathbf{V}=\left( \mathbf{V}_{1},\ldots ,\mathbf{V}_{N}\right) \in {%
\mathbb{R}}^{6N}$ comprises the twist vectors $\mathbf{V}_{i}=\left( 
\bm{\omega}%
_{i},\mathbf{v}_{i}\right) $ of all $N$ bodies relative to a global inertial
frame, and the coordinate vector $\mathbf{q}=\left( \mathbf{q}_{1},\ldots ,%
\mathbf{q}_{N}\right) \in {\mathbb{V}}^{n},n\geq 6N$ consists of the vector
of absolute coordinates $\mathbf{q}_{i}\in {\mathbb{V}}^{n_{i}},n_{i}\geq 6$
used to describe the pose of body $i$. If non-redundant coordinates are
used, then $n_{i}=6,n=6N$. Notice that, strictly speaking, $\mathbf{q}_{i}$
do not represent coordinates when $n_{i}>6$, i.e. when they are dependent as
in case of unit quaternions, where $n_{i}=7$. However, also for $n_{i}>6$,
the dependent parameters $\mathbf{q}_{i}$ are referred to as absolute
coordinates in the literature, which also adopted throughout the paper. The
system (\ref{GeomConst}) of $m$ equations represents (for simplicity
scleronomic) geometric constraints. Along with further non-holonomic
constraints, they give rise to a system of $\bar{m}$ velocity constraints (%
\ref{VelConst}). The vector $%
\bm{\lambda}%
$ of Lagrange multipliers accounts for the constraint reactions. The EOM (%
\ref{EOMDyn}) govern the dynamics, and the kinematics EOM (\ref{EOMkin})
relate the state of the MBS to the absolute coordinates. The equations (\ref%
{EOMDyn})-(\ref{GeomConst}) are transformed to the system%
\begin{eqnarray}
\left( 
\begin{array}{cc}
\mathbf{M}\left( \mathbf{q}\right)  & \mathbf{A}^{T} \\ 
\mathbf{A} & \mathbf{0}%
\end{array}%
\right) \left( 
\begin{array}{c}
\dot{\mathbf{V}} \\ 
\bm{\lambda}%
\end{array}%
\right)  &=&\left( 
\begin{array}{c}
\mathbf{Q}\left( \mathbf{q},\mathbf{V},t\right)  \\ 
-\dot{\mathbf{A}}\mathbf{V}%
\end{array}%
\right)   \label{EOM1} \\
\mathbf{V} &=&\mathbf{H}\left( \mathbf{q}\right) \dot{\mathbf{q}}.
\label{EOM2}
\end{eqnarray}%
The equations (\ref{EOM1}),(\ref{EOM2}) can be solved for the time
derivative of the state $\left( \mathbf{q},\mathbf{V}\right) $ to provide
the following index 1 DAE system with hidden constraints (\ref{GeomConst})%
\begin{eqnarray}
\dot{\mathbf{V}} &=&f\left( \mathbf{q},\mathbf{V},t\right)   \label{EOMdyn2}
\\
\dot{\mathbf{q}} &=&\mathbf{H}^{-1}\left( \mathbf{q}\right) \mathbf{V}.
\label{EOMkin2}
\end{eqnarray}%
The formulation (\ref{EOM1}),(\ref{EOM2}) is frequently used and integrated
with standard ODE schemes in conjunction with constraint stabilization
methods \cite{NikraveshBook1988}.

The EOM govern the dynamics and kinematics of the MBS evolving on the
admissible parameter space $V:=\{\mathbf{q}\in {\mathbb{V}}^{n}|g\left( 
\mathbf{q}\right) =\mathbf{0}\}\subset {\mathbb{V}}^{n}$ defined by the
constraints (\ref{GeomConst}). Geometrically, however, the configuration
space of the MBS is a subspace of a Lie group $G$, while ${\mathbb{V}}^{n}$
is the parameter space corresponding to the parameterization of $G$ with
absolute coordinates. The geometry of the state space leads to two principal
challenges for numerical time integration: 1) the configuration space $V$ is
embedded in a higher-dimensional manifold ${\mathbb{V}}^{n}$, and 2) the
absolute coordinates used in the MBS model correspond to a parameterization
of the Lie group $G$. In practical terms, the first issue has to do with the
satisfaction of geometric constraints, and the second with the
parameterization of the MBS model. Various methods for stabilizing
constraint violations during numerical integration were proposed, including
the projection methods \cite{Blajer2001,Brauchli1991,TerzeNaudet2008}. The
second issue, i.e. parameterizing rigid body motions, is a classical topic,
and the principal problem is the singularity-free and geometrically
consistent description spatial rotations. The standard approach to avoid
singularities is to use unit quaternions (Euler parameters), which represent
redundant coordinates, thus $n>6N$. While this enables a computationally
simple algebraic description, it involves four dependent rotation parameters
per body, and the corresponding unit norm constraint must be satisfied
during numerical integration. Common approaches to enforce the unit norm
constraint are the renormalization in each integration step \cite{Barker1973}%
, or to include them within the overall system of constraints \cite%
{Nikravesh1985-1,Shabana2014}, which are enforced via Lagrange multipliers,
but there are also more advanced approaches \cite{MollerGlocker2012}. This
is known to interfere with the system dynamics, however \cite%
{Barker1973,Nikravesh1985-1,Shabana2014,TerzeMueller2016,HaghshenasJaryaniBowling2013}%
. Among other benefits, Lie group methods allow addressing both issues. They
give rise to singularity-free descriptions (by using local coordinates
within integration steps) and allow to determine the geometrically correct
finite motion within an integration step \cite%
{BorriTrainelliBottasso2000,MMTCSpace2014,BIT2016,BIT2016Erratum,SonnevilleCardonaBruls2014}%
. The latter, for instance, improves constraint satisfaction (by respecting
isotropy groups defined by technical joints, i.e. they restrict relative
motions to the contact surface of lower pairs, which is invariant under the
joint motion \cite{BIT2016} when modeled on $SE\left( 3\right) $).

\section{Equations of Motion in Lie Group Descriptor Form%
\label{secEOMLie}%
}

\subsection{Configuration and State Space Lie Group}

The MBS comprises $N$ rigid bodies. The absolute configuration (pose) of
body $i$ relative to a global inertial frame is represented by $C_{i}\in
G_{i}$, where $G_{i}$ is a Lie group. The MBS configuration is represented
as $C=\left( C_{1},\ldots ,C_{N}\right) \in G$, where $G=G_{1}\times \ldots
\times G_{N}$ is the ambient Lie group of the configuration space. The
configuration space is the subspace of $G$ defined by the system of
geometric constraints denoted $g\left( C\right) =\mathbf{0}$ (with slight
abuse of notation of (\ref{GeomConst})), which in general, is not a Lie
group itself.

Absolute coordinates provide a parameterization of the configuration of body 
$i$ as $C_{i}=\alpha _{i}\left( \mathbf{q}_{i}\right) $, and thus of the
MBS, $C=\alpha \left( \mathbf{q}\right) $, where the choice of absolute
coordinates corresponds to a coordinate map 
\begin{equation}
\alpha :\mathbf{q}\in {\mathbb{V}}^{\nu }\mapsto \alpha \left( \mathbf{q}%
\right) \in G.
\end{equation}%
This parameterization is usually not global. It is globally valid only if
unit quaternions are used to describe rotations.

The velocity of the MBS is defined by left-trivialization%
\begin{equation}
\mathbf{V}=\left( \mathbf{V}_{1},\ldots ,\mathbf{V}_{N}\right)
=T_{C}L_{C^{-1}}\dot{C}=C^{-1}\dot{C}\in \mathfrak{g}.  \label{KinRecLeft}
\end{equation}%
Relation (\ref{KinRecLeft}) is the generalized (left) Poisson-Darboux
equation on $G$ \cite{ConduracheAASAIAA2017,RSPA2021}. It is referred to as
the \emph{kinematic reconstruction equation} as it relates the velocity $%
\mathbf{V}\left( t\right) \in \mathfrak{g}$ of the system to its actual
motion $C\left( t\right) \in G$. The state space of the MBS is the subspace
of $G\times \mathfrak{g}\cong G\times {\mathbb{R}}^{6N}$ defined by the $m$
geometric constraints and the $\bar{m}$ velocity constraints.

\begin{remark}
The MBS velocity can also be defined by right-trivialization%
\begin{equation}
\hat{\mathbf{V}}=\left( \hat{\mathbf{V}}_{1},\ldots ,\hat{\mathbf{V}}%
_{N}\right) =T_{C}R_{C^{-1}}\dot{C}=\dot{C}C^{-1}.  \label{KinRecRight}
\end{equation}%
If $G_{i}=SE\left( 3\right) $, then $\mathbf{V}_{i}\in se\left( 3\right) $
is the twist of body $i$ in spatial representation determined by $\hat{%
\mathbf{V}}_{i}=\dot{\mathbf{C}}_{i}\mathbf{C}_{i}^{-1}\in se\left( 3\right) 
$ \cite{MUBOScrew1}, which is also called the fixed-pole representation \cite%
{Bottasso1998,Borri2001b}. If $G_{i}=SO\left( 3\right) \times {\mathbb{R}}%
^{3}$, then $\mathbf{V}_{i}\in so\left( 3\right) \times {\mathbb{R}}^{3}$ is
the twist in hybrid representation \cite{MUBOScrew1}. Throughout this paper,
the body-fixed and mixed representation of twists are used, as they prevail
in MBS dynamics.
\end{remark}

\subsection{DAE Formulation}

The EOM of a constrained MBS in Lie group descriptor form can be expressed
by the DAE system%
\begin{eqnarray}
\mathbf{M}\dot{\mathbf{V}}+\mathbf{A}^{T}\left( C\right) 
\bm{\lambda}
&=&\mathbf{Q}\left( C,\mathbf{V},t\right)  \label{LieEOMDyn} \\
\dot{C} &=&T_{I}L_{C}\cdot \mathbf{V}  \label{LieEOMKin} \\
\mathbf{A}\left( C\right) \mathbf{V} &=&\mathbf{0}  \label{LieVelConst} \\
\mathbf{g}\left( C\right) &=&\mathbf{0}  \label{LieGeomConst}
\end{eqnarray}%
where (\ref{LieEOMKin}) accounts for the left-trivialized representation of
velocities replacing (\ref{EOMkin}), respectively (\ref{EOMkin2}), and $I$
is the identity in $G$. The system mass matrix $\mathbf{M}$ is constant as
the kinetic energy is expressed in terms of left-invariant (w.r.t. actions
of $G$) velocities. In Lie group context, equations (\ref{LieEOMDyn}) can be
regarded as the Euler-Lagrange equations for the left-trivialized Lagrangian
of the rigid body MBS.

In analogy to (\ref{EOMdyn2}), the equations (\ref{LieEOMDyn})-(\ref%
{LieVelConst}) can be resolved as%
\begin{eqnarray}
\dot{\mathbf{V}} &=&f\left( C,\mathbf{V},t\right)  \label{LieEOMDyn2} \\
\dot{C} &=&T_{I}L_{C}\cdot \mathbf{V}  \label{LieEOMKin2}
\end{eqnarray}%
which is an index 1 DAE on $G\times \mathfrak{g}\cong G\times {\mathbb{R}}%
^{6N}$ with hidden constraints (\ref{LieGeomConst}). More precisely, (\ref%
{LieEOMDyn2}) is a DAE system on the vector space ${\mathbb{R}}^{6N}$, and (%
\ref{LieEOMKin2}) is an ODE system on the Lie group $G$. Of course, the
vector space DAE formulation in sec. \ref{secVecDAE} is recovered by
introducing an absolute coordinate parameterization in the above Lie group
formulation. Throughout the paper, and without loss of generality of the
presented concept, DAEs of the form (\ref{LieEOMDyn2}),(\ref{LieEOMKin2})
are considered.

\section{Lie Group Integration Methods for Constrained MBS%
\label{secLieInt}%
}

The above Lie group DAE formulation serves as representative for a wide
class of Lie group formulations for mechanical systems. They can be solved
with numerical integration schemes operating directly on the Lie group $%
G\times {\mathbb{R}}^{6N}$. An excellent overview on Lie group integrators
for ODE systems can be found in \cite%
{IserlesMuntheKaasNrsettZanna2000,CelledoniOwren2003,Owren2018}. Multistep
Runge-Kutta methods were introduced by Munthe-Kaas in \cite%
{MuntheKaas1998,MuntheKaas1999}. Based on these methods, a multistep BDF
scheme on Lie groups, called BLieDF methods, was presented in \cite%
{WielochArnold2021}. For DAE systems, describing multibody system models in
absolute coordinates, the established Newmark and generalized-$\alpha $
schemes \cite{Newmark1959} were adopted to the Lie group formulation (\ref%
{LieEOMDyn})-(\ref{LieVelConst}) in \cite%
{BrulsCardonaArnold2012,BrulsCardona2010,ArnoldHante2017,ArnoldBrulsCardona2015}%
. Yet, any such method employs local coordinates for the update step within
the integration scheme, so that they cannot be used along with the classical
absolute coordinate models (\ref{EOMDyn})-(\ref{GeomConst}) or (\ref{EOMdyn2}%
),(\ref{EOMkin2}). The Munthe-Kaas methods \cite%
{MuntheKaas1998,MuntheKaas1999}, for example, use canonical coordinates on $G
$ along with the exp map for computing the update step.

\subsection{Local Parameterization}

The DAE system (\ref{LieEOMDyn2}),(\ref{LieEOMKin2}) and (\ref{EOMDyn})-(\ref%
{GeomConst}) is defined on the Lie group $G\times {\mathbb{R}}^{6N}$. The
vector space ${\mathbb{R}}^{6N}$ possesses an obvious coordinate chart
defined by the frames in which twists are represented. Various local
coordinates can be introduced on $G$ in order to solve the kinematic
reconstruction equations (\ref{LieEOMKin2}). Let $\psi :\mathfrak{g}%
\rightarrow G$ be a (local) coordinate map on $G$. The solution of (\ref%
{LieEOMKin}) can be expressed as%
\begin{equation}
C\left( t\right) =C\left( 0\right) \psi \left( \mathbf{X}\left( t\right)
\right)  \label{Ct}
\end{equation}%
where $\mathbf{X}$ is a curve in $\mathfrak{g}$, and $\mathbf{X}\left(
t\right) =\left( \mathbf{X}_{1},\ldots ,\mathbf{X}_{N}\right) \in \mathfrak{g%
}$ is the vector of (local) coordinates on $G$. More precisely, $\mathbf{X}%
_{i}\in \mathfrak{g}_{i}$ are local coordinates on $G_{i}$, and the solution
for body $i$ is $C_{i}\left( t\right) =C_{i}\left( 0\right) \psi _{i}\left( 
\mathbf{X}_{i}\left( t\right) \right) $ (notice left multiplication since
body-fixed twists are used in (\ref{LieEOMKin})). This use of local
coordinates is central to all Lie group methods. As an important
representative, Munthe-Kaas methods make use of the corresponding
differential relation as follows.

If (\ref{Ct}) is a solution of (\ref{LieEOMKin}), it satisfies the ODE on $%
\mathfrak{g}_{i}$ \cite{Iserles1984}%
\begin{equation}
{\mathbf{V}}=\mathrm{d}\psi _{-\mathbf{X}}\left( \dot{\mathbf{X}}\right) ,
\label{dpsi}
\end{equation}%
where $\mathrm{d}\psi _{\mathbf{X}}\left( \mathbf{Y}\right) $ is the
right-trivialized differential of $\psi $, defined by $\left( D_{\mathbf{X}%
}\psi \right) \left( \mathbf{Y}\right) =\mathrm{d}\psi _{\mathbf{X}}\left( 
\mathbf{Y}\right) \psi \left( \mathbf{X}\right) ,\mathbf{X},\mathbf{Y}\in 
\mathfrak{g}_{i}$, and $D_{\mathbf{X}}\psi $ is the directional derivative
of $\psi $ along $\mathbf{X}$.

The kinematic reconstruction equations (\ref{LieEOMKin2}) are replaced by (%
\ref{dpsi}), and the overall system of EOM can be written as an index 2 DAE
system on the ambient state (vector) space $\mathfrak{g}\times {\mathbb{R}}%
^{6N}\cong {\mathbb{R}}^{6N}\times {\mathbb{R}}^{6N}$%
\begin{eqnarray}
\mathbf{M}\dot{\mathbf{V}}+\mathbf{A}^{T}\left( C\right) 
\bm{\lambda}
&=&\mathbf{Q}\left( C,\mathbf{V},t\right)  \label{LieEOMDyn5} \\
\dot{\mathbf{X}} &=&\mathrm{d}\psi _{-\mathbf{X}}^{-1}\left( {\mathbf{V}}%
\right)  \label{LieEOMKin5} \\
\mathbf{A}\left( C\right) \mathbf{V} &=&\mathbf{0} \\
\mathbf{g}\left( C\right) &=&\mathbf{0}  \label{LieGeomConst5}
\end{eqnarray}%
or as an index 1 DAE system 
\begin{eqnarray}
\dot{\mathbf{V}} &=&\varphi \left( C,\mathbf{V},t\right)  \label{LieEOMDyn3}
\\
\dot{\mathbf{X}} &=&\mathrm{d}\psi _{-\mathbf{X}}^{-1}\left( {\mathbf{V}}%
\right)  \label{LieEOMKin3}
\end{eqnarray}%
with $C\left( t\right) =C\left( \mathbf{X}\left( t\right) \right) $ in (\ref%
{Ct}). The kinematic equations (\ref{LieEOMKin3}) on $G$ are intrinsic to
the Lie group used for representing rigid body motions, and the local
coordinates on $G$.%
\newpage%

\subsection{Integration Schemes}

\paragraph{Munthe-Kaas Method:}

Replacing the kinematic reconstruction equations (\ref{LieEOMKin}) by (\ref%
{dpsi}) is the starting point for many Lie group methods. The system
dynamics is now evolving on a vector space, which allows for time
integration with standard vector space integration schemes. A typical
approach is to solve the system%
\begin{eqnarray}
\dot{\mathbf{V}} &=&\varphi \left( C,\mathbf{V},t\right)   \label{LieEOMDyn4}
\\
\dot{\mathbf{X}} &=&\mathrm{d}\psi _{-\mathbf{X}}^{-1}\left( {\mathbf{V}}%
\right)   \label{dpsiinv} \\
C\left( t\right)  &=&C\left( t_{k}\right) \psi \left( \mathbf{X}\left(
t\right) \right)   \label{Step}
\end{eqnarray}%
in time step $k+1$, within the time interval $t\in \left[ t_{k},t_{k+1}%
\right] $, with initial value $\mathbf{X}\left( t_{k}\right) =\mathbf{0}$,
and with $C\left( t_{k}\right) $ given from the previous time step. Using
the original Munthe-Kaas method \cite{MuntheKaas1998} or multistep methods 
\cite{FaltinsenMarthinsenMuntheKaas2001}, the system (\ref{LieEOMDyn4})-(\ref%
{Step}) is solved with a Runge-Kutta scheme, and the $\exp $ mapping is used
as local coordinate map $\psi $. Constraint stabilization methods must be
applied since (\ref{LieEOMDyn4}) is index 1 DAE system, however. An
important aspect is the convergence analysis of the multistep methods, and
when $\mathrm{dexp}^{-1}$ is approximated by a truncated series, the
reduction of the number of Lie brackets to be evaluated is also an issue 
\cite{MuntheKaas1999,IserlesMuntheKaasNrsettZanna2000}.

\paragraph{Generalized-$\protect\alpha $ and BDF Methods:}

There are two relevant Lie group integration schemes for constrained MBS
that were proposed recently. In \cite{WielochArnold2021}, a multistep BDF
scheme was presented. The method, called BLieDF is optimized so to reduce
the number of Lie bracket computations. Another method is the generalized-$%
\alpha $ method that was proposed for integration of the index 3 EOM in Lie
group descriptor form \cite%
{BrulsCardonaArnold2012,BrulsCardona2010,ArnoldHante2017,ArnoldBrulsCardona2015}%
. Both methods do not make explicit use of the kinematic equations (\ref%
{dpsiinv}), but the differential is central for the derivation of the
schemes.

While the algorithm-specific integration steps are different for different
Lie group methods, any such method requires evaluation of the EOM in terms
of the Lie group element $C\in G$, rather than in terms of absolute
coordinates $\mathbf{q}$. Lie group methods use a coordinate free
representation of rigid body motions, and local coordinates on $G$ are only
used within the integration step. The update of the global configuration is
performed with (\ref{Step}) (an exception is the Crouch-Grossman method \cite%
{ChrouchGrossman1993}, where a product of exponentials is used in (\ref{Step}%
)). It should be emphasized that the coordinate map $\psi $ only needs to
provide a local parameterization. This is the reason why Lie group methods
allow for singularity-free integration even if there is no singularity-free
global parameterization, as for $SO\left( 3\right) $ in case of spatial
rotations. In this sense, (\ref{dpsiinv}) can be regarded as \emph{local
kinematic reconstruction equations}. Any canonical or non-canonical local
coordinates can be used \cite{OwrenMarthinsen2001}.

Since in the past, absolute coordinate models have been developed for many
complex MBS, it is most natural aiming to make Lie group integrations
schemes applicable to such models. That is, when the classical formulations (%
\ref{EOMDyn})-(\ref{GeomConst}) or (\ref{EOMdyn2}),(\ref{EOMkin2}) of the
EOM are to be solved with Lie group methods, the absolute coordinates $%
\mathbf{q}\left( t\right) ,t\in \left[ t_{k},t_{k+1}\right] $ must be
computed from the absolute coordinates $\mathbf{q}\left( t_{k}\right) $ and
local coordinates $\mathbf{X}\left( t\right) $ during the integration step $%
t_{k+1}$. This is addressed in the remainder of this paper.

\section{Relevant Lie Groups and Coordinate Maps%
\label{secCoordMaps}%
}

Local coordinate maps $\psi $ are used within the update step (\ref{Step})
of the Lie group integration scheme. They are specific to the Lie group used
to represent rigid body motions. The most relevant Lie groups and local
coordinates are summarized in this section.

\subsection{Direct product Lie group $SO\left( 3\right) \times {\mathbb{R}}%
^{3}$}

The pose of a rigid body relative to the IFR can be represented by $C=\left( 
\mathbf{R},\mathbf{r}\right) \in SO\left( 3\right) \times {\mathbb{R}}^{3}$,
where the rotation matrix $\mathbf{R}\in SO\left( 3\right) $ transforms
coordinates related to the body-fixed frame to those related to the IFR, and 
$\mathbf{r}\in {\mathbb{R}}^{3}$ is the position vector of the body-fixed
frame expressed in the IFR. Generally, $C$ describes the relative
configuration of two frames. Multiplication on the direct product group is
defined as $C_{1}\cdot C_{2}=\left( \mathbf{R}_{1}\mathbf{R}_{2},\mathbf{r}%
_{1}+\mathbf{r}_{2}\right) $. Describing rigid body configurations with the
direct product $SO\left( 3\right) \times {\mathbb{R}}^{3}$ implies that
rotations and translations are decoupled, and the coordinate maps are given
by those on the two factors. Thus, $G_{i}:=SO\left( 3\right) \times {\mathbb{%
R}}^{3}$ does represent rigid body configurations, but cannot capture the
geometry of rigid body motions, i.e. successive application of frame
transformations. This representation is motivated by classical MBS\
formulations, where the kinematic motion equations for position and attitude
are solved separately (i.e. $\mathbf{v}_{i}=\dot{\mathbf{r}}_{i}$ and $\dot{%
\mathbf{x}}_{i}=\mathbf{dexp}_{-\mathbf{x}_{i}}^{-1}%
\bm{\omega }%
$, see section \ref{secExpSO3xR3}), which stems from the fact that the
dynamic Newton-Euler equations are decoupled when expressed w.r.t. the
center of mass (COM), and thus the translation of the COM and the rotation
about the COM are decoupled. This is not the case when the body-fixed
reference frames are generally located off the COM, and for constrained MBS
in general \cite{MMTCSpace2014,BIT2016}.

\paragraph{Exponential Map%
\label{secExpSO3xR3}%
}

The exp map $\psi :=\exp $ is a typical choice for parameterization with
canonical coordinates (of first kind). On the direct product, the exp map is 
\begin{equation}
\mathbf{X}=\left( \mathbf{x},\mathbf{r}\right) \in {\mathbb{R}}^{6}\mapsto
\exp (\mathbf{X})=\left( \exp \tilde{\mathbf{x}},\mathbf{r}\right) \in
SO\left( 3\right) \times {\mathbb{R}}^{3}  \label{expSO3R3}
\end{equation}%
where the exp map on $SO\left( 3\right) $ describes a frame rotation.

Denote with $\mathbf{n}\in {\mathbb{R}}^{3}$ a unit vector along the
rotation axis, with $\varphi $ the rotation angle, then $\mathbf{x}=\varphi 
\mathbf{n}$ is the scaled rotation axis. The rotation matrix describing the
corresponding frame rotation is then $\mathbf{R}=\exp \tilde{\mathbf{x}}%
=\exp \left( \varphi \tilde{\mathbf{n}}\right) $, where the exp map on $%
SO\left( 3\right) $ possesses the closed form expressions \cite%
{BorriTrainelliBottasso2000,RSPA2021}%
\begin{eqnarray}
\exp \tilde{\mathbf{x}} &=&\mathbf{I}+\alpha \tilde{\mathbf{x}}+\tfrac{1}{2}%
\beta \tilde{\mathbf{x}}^{2}  \label{SO3exp1} \\
\exp \left( \varphi \tilde{\mathbf{n}}\right) &=&\mathbf{I}+\sin \varphi \,%
\tilde{\mathbf{n}}+\tfrac{1}{2}\sin ^{2}\frac{\varphi }{2}\,\tilde{\mathbf{n}%
}^{2}  \label{SO3exp2}
\end{eqnarray}%
with $\alpha :=\mathrm{sinc}\varphi ,\beta :=\mathrm{sinc}^{2}\mathrm{\,}%
\frac{\varphi }{2}$, where $\mathrm{sinc}\varphi =\left( \sin \varphi
\right) /\varphi $.

The time derivative of $\mathbf{x}$ determines the body-fixed angular
velocity as $%
\bm{\omega }%
=\mathbf{dexp}_{-\mathbf{x}}\dot{\mathbf{x}}$, with the right-trivialized
differential%
\begin{equation}
\mathbf{dexp}_{\mathbf{x}}=\mathbf{I}+\tfrac{\beta }{2}\tilde{\mathbf{x}}%
+\left( 1-\alpha \right) \tilde{\mathbf{n}}^{2}  \label{dexpSO3}
\end{equation}%
In the inverse relation $\dot{\mathbf{x}}=\mathbf{dexp}_{-\mathbf{x}}^{-1}%
\bm{\omega }%
$, the closed form relation%
\begin{equation}
\mathbf{dexp}_{\mathbf{x}}^{-1}=\mathbf{I}-\tfrac{1}{2}\tilde{\mathbf{x}}+%
\tfrac{1}{\left\Vert \mathbf{x}\right\Vert ^{2}}\left( 1-\gamma \right) 
\tilde{\mathbf{x}}^{2}  \label{SO3dexpInv}
\end{equation}%
is used with $\gamma :=\alpha /\beta $.

When representing motions with $SO\left( 3\right) \times {\mathbb{R}}^{3}$,
then $\hat{\mathbf{V}}_{i}=C_{i}^{-1}\dot{C}_{i}\in so\left( 3\right) \times 
{\mathbb{R}}^{3}$ is the body velocity in mixed representation, which yields
the velocity vector $\mathbf{V}_{i}=\left( {^{i}}%
\bm{\omega}%
_{i},\mathbf{v}_{i}\right) \in {\mathbb{R}}^{6}$, where ${^{i}}%
\bm{\omega}%
_{i}$ is the angular velocity represented in body-fixed frame and $\mathbf{v}%
_{i}=\dot{\mathbf{r}}_{i}$ is the linear velocity in IFR \cite%
{MUBOScrew1,Murray} (the leading superscript in ${^{i}}%
\bm{\omega}%
_{i}$ and $\mathbf{v}_{i}$ indicate that the vectors are resolved in the
body-frame on body $i$). The differential and relation (\ref{dpsi}) is
simply $\mathbf{V}=\left( \mathbf{dexp}_{-\mathbf{x}}\dot{\mathbf{x}},\dot{%
\mathbf{r}}\right) $\textbf{. }The latter is given as $\mathbf{V}=\mathbf{%
dexp}_{-\mathbf{x}}\dot{\mathbf{X}}$ in terms of the right-trivialized
differential of the exp map on $SO\left( 3\right) \times {\mathbb{R}}^{3}$
in matrix form 
\begin{equation}
\mathbf{dexp}_{\mathbf{X}}=\left( 
\begin{array}{cc}
\mathbf{dexp}_{\mathbf{x}} & \ \ \mathbf{0} \\ 
\mathbf{0} & \mathbf{I}%
\end{array}%
\right) .  \label{dexpSO3xR3}
\end{equation}

\paragraph{Cayley Map}

A typical choice for non-canonical parameterization is the Cayley map $\psi
:=\mathrm{cay}$. On $SO\left( 3\right) $ it possesses the explicit form \cite%
{BorriTrainelliBottasso2000,RSPA2021,Selig-IFToMM2007}%
\begin{equation}
\mathrm{cay}\left( \tilde{\mathbf{c}}\right) =\mathbf{I}+\sigma \left( 
\tilde{\mathbf{c}}+\tilde{\mathbf{c}}^{2}\right)  \label{CaySO31}
\end{equation}%
where $\sigma =2/(1+\left\Vert \mathbf{c}\right\Vert ^{2})$. The vector $%
\mathbf{c}\in {\mathbb{R}}^{3}$ is called Gibbs-Rodrigues vector, while its
components are referred to as the Rodrigues parameters. The
right-trivialized differential and its inverse are%
\begin{eqnarray}
\mathbf{dcay}_{\mathbf{c}} &=&\sigma \left( \mathbf{I}+\tilde{\mathbf{c}}%
\right)  \label{dCaySO3} \\
\mathbf{dcay}_{\tilde{\mathbf{c}}}^{-1} &=&\frac{1}{2\sigma }\left( \mathbf{I%
}+\mathrm{cay}\left( -\tilde{\mathbf{c}}\right) \right) =\frac{1}{\sigma }%
\mathbf{I}+\frac{1}{2}(\tilde{\mathbf{c}}^{2}-\tilde{\mathbf{c}})
\label{dCayInvSO32}
\end{eqnarray}%
which satisfy $\dot{\mathbf{c}}=\mathbf{dcay}_{\tilde{\mathbf{c}}}^{-T}%
\bm{\omega }%
$ and $%
\bm{\omega }%
=\mathbf{dcay}_{\tilde{\mathbf{c}}}^{T}\dot{\mathbf{c}}$, respectively. The
body velocity is $\mathbf{V}=\left( \mathbf{dcay}_{\tilde{\mathbf{c}}}^{T}%
\dot{\mathbf{c}},\dot{\mathbf{r}}\right) $\textbf{. }

\subsection{Direct product Lie group $Sp\left( 1\right) \times {\mathbb{R}}%
^{3}$, Unit Quaternions}

The direct product $Sp\left( 1\right) \times {\mathbb{R}}^{3}$ is
homomorphic to $SO\left( 3\right) \times {\mathbb{R}}^{3}$, and is relevant
when unit quaternions are used as absolute coordinates. A typical element
has the form $\left( \mathbf{Q},\mathbf{r}\right) \in Sp\left( 1\right)
\times {\mathbb{R}}^{3}$, where $\mathbf{Q}\in Sp\left( 1\right) $ is a unit
quaternion representing the orientation of a body, and $\mathbf{r}\in {%
\mathbb{R}}^{3}$ is a position vector. It serves as configuration space Lie
group, but at the same time, $\left( \mathbf{Q},\mathbf{r}\right) $ can be
used as redundant absolute coordinates.

A unit quaternion is a four tuple of the form $\mathbf{Q}=\left( p_{0},%
\mathbf{p}\right) $ with scalar (also called real) part $p_{0}\in {\mathbb{R}%
}$, and vector (also called complex) part $\mathbf{p}=\left(
p_{1},p_{2},p_{3}\right) \in {\mathbb{R}}^{3}$ satisfying the condition $%
\left\Vert \mathbf{Q}\right\Vert
=p_{0}^{2}+p_{1}^{2}+p_{2}^{2}+p_{3}^{2}=p_{0}^{2}+\left\Vert \mathbf{p}%
\right\Vert =1$. Multiplication of two unit quaternions $\mathbf{Q}%
_{1}=\left( q_{10},\mathbf{q}_{1}\right) $ and $\mathbf{Q}_{2}=\left( q_{20},%
\mathbf{q}_{2}\right) $ is defined as 
\begin{equation}
Q_{1}\cdot Q_{2}=\left( q_{10}q_{20}-\mathbf{q}_{1}\cdot \mathbf{q}%
_{2},q_{10}\mathbf{q}_{2}+q_{20}\mathbf{q}_{1}+\mathbf{q}_{1}\times \mathbf{q%
}_{2}\right) .  \label{Qmult}
\end{equation}

Unit quaternions form the Lie group $Sp\left( 1\right) $, which is
homomorphic to $SO\left( 3\right) $ (to each rotation matrix $\mathbf{R}\in
SO\left( 3\right) $ correspond two unit quaternions $\mathbf{Q}\in Sp\left(
1\right) $). Thus, both have the same canonical coordinates, and the unit
quaternion describing the rotation about axis $\mathbf{x}$ and angle $%
\varphi $ is determined by the exponential map $\exp :sp\left( 1\right)
\mapsto Sp\left( 1\right) $ \cite{CND2016}%
\begin{eqnarray}
\mathbf{Q} &=&\exp (0,\frac{1}{2}\mathbf{x})  \notag \\
&=&(\cos \tfrac{\left\Vert \mathbf{x}\right\Vert }{2},\tfrac{1}{2}\mathrm{%
sinc}\tfrac{\left\Vert \mathbf{x}\right\Vert }{2}\mathbf{x})=(\cos \tfrac{%
\varphi }{2},\mathbf{n}\sin \tfrac{\varphi }{2}).  \label{Sp1exp}
\end{eqnarray}%
The corresponding rotation matrix $\mathbf{R}=\exp \tilde{\mathbf{x}}=\exp
\left( \varphi \tilde{\mathbf{n}}\right) $ can be expressed in terms of a
unit quaternion as%
\begin{equation}
\mathbf{R}=\mathbf{I}+2\left( p_{0}\widetilde{\mathbf{p}}+\widetilde{\mathbf{%
p}}^{2}\right) .  \label{QuatR}
\end{equation}%
Clearly, $\mathbf{Q}$ and $-\mathbf{Q}$ yield the same ration matrix.
Moreover, the Lie algebras $sp\left( 1\right) $ and $so\left( 3\right) $ are
isomorphic, $sp\left( 1\right) \cong so\left( 3\right) \cong {\mathbb{R}}%
^{3} $, so that the differential of the exp maps are identical to (\ref%
{dexpSO3}), and the canonical coordinates in (\ref{Sp1exp}) satisfy $%
\bm{\omega }%
=\mathbf{dexp}_{-\mathbf{x}}\dot{\mathbf{x}}$, with $\mathbf{dexp}$ in (\ref%
{dexpSO3}).

A unit quaternion can be expressed in terms of a Gibbs-Rodrigues vector $%
\mathbf{c}\in {\mathbb{R}}^{3}$ as \cite{CND2016}%
\begin{equation}
p_{0}=\sqrt{\sigma /2},\ \mathbf{p}=\sqrt{1-\sigma /2}~\mathbf{c}.
\label{QuatRodrigues}
\end{equation}

The direct product $Sp\left( 1\right) \times {\mathbb{R}}^{3}$ implies
decoupled rotations and translations as above. Thus, the parameterization
with canonical coordinates via the exponential map is 
\begin{equation}
\mathbf{X}=\left( \mathbf{x},\mathbf{r}\right) \in {\mathbb{R}}^{3}\times {%
\mathbb{R}}^{3}\mapsto \exp (\mathbf{X})=\left( \exp (0,\tfrac{1}{2}\mathbf{x%
}),\mathbf{r}\right) \in Sp\left( 1\right) \times {\mathbb{R}}^{3}.
\label{expSp1}
\end{equation}

\subsection{Special Euclidean Group $SE\left( 3\right) $}

The semidirect product Lie group $SE\left( 3\right) =SO\left( 3\right)
\ltimes {\mathbb{R}}^{3}$ also consists of elements $C=\left( \mathbf{R},%
\mathbf{r}\right) \in SE\left( 3\right) $, but is equipped with the
multiplication $C_{1}\cdot C_{2}=\left( \mathbf{R}_{1}\mathbf{R}_{2},\mathbf{%
r}_{1}+\mathbf{R}_{1}\mathbf{r}_{2}\right) $. Now $G_{i}:=SE\left( 3\right) $
clearly describes frame transformations, and thus proper rigid body motions.
Frame transformations are conveniently expressed with homogenous
transformation matrices of the form%
\begin{equation}
\mathbf{C}=\left( 
\begin{array}{cc}
\mathbf{R} & \mathbf{r} \\ 
\mathbf{0} & 1%
\end{array}%
\right) \in SE\left( 3\right) .  \label{C}
\end{equation}%
Then, the multiplication $C_{1}\cdot C_{2}$ becomes the matrix
multiplication $\mathbf{C}_{1}\mathbf{C}_{2}$. Clearly, the semidirect
product $SE\left( 3\right) =SO\left( 3\right) \ltimes {\mathbb{R}}^{3}$
describes proper frame transformation, i.e. rigid body motions, with coupled
rotations and translations. This observation is crucial for modeling and
simulation of spatial MBS \cite%
{MMTCSpace2014,BIT2016,BIT2016Erratum,SonnevilleCardonaBruls2014,BorriTrainelliBottasso2000}%
.

Dual quaternions form the Lie group $\widehat{Sp}\left( 1\right) $ which is
homomorphic to $SE\left( 3\right) $. As dual quaternions are not used in MBS
modeling, this will not be considered.

\paragraph{Exponential Map}

A rigid body motion is a screw motion, and screw coordinates, denoted $%
\mathbf{X}=\left( \mathbf{x},\mathbf{y}\right) \in {\mathbb{R}}^{6}\cong
se\left( 3\right) $, are canonical coordinates on $SE\left( 3\right) $. The
exp map is given explicitly in terms of screw coordinates \cite%
{BorriTrainelliBottasso2000,RSPA2021}%
\begin{eqnarray}
\exp (\mathbf{X}) &=&\left( 
\begin{array}{cc}
\exp \tilde{\mathbf{x}} & \ \ \mathbf{dexp}_{\mathbf{x}}\mathbf{y} \\ 
\mathbf{0} & 1%
\end{array}%
\right)   \label{expSE31} \\
&=&\left\{ 
\begin{array}{l}
\left( 
\begin{array}{cc}
\mathbf{R} & \ \ \left( \mathbf{I}+\tfrac{\beta }{2}\tilde{\mathbf{x}}%
+\left( 1-\alpha \right) \tilde{\mathbf{n}}^{2}\right) \mathbf{y} \\ 
\mathbf{0} & 1%
\end{array}%
\right) ,\ \mathrm{for\ }\mathbf{x}\neq \mathbf{0}%
\vspace{1ex}
\\ 
\left( 
\begin{array}{cc}
\mathbf{I} & \ \mathbf{y} \\ 
\mathbf{0} & 1%
\end{array}%
\right) ,\ \mathrm{for\ }\mathbf{x}=\mathbf{0}%
\end{array}%
\right. \ \ \ \ \ \ \   \label{expSE32}
\end{eqnarray}%
with $\mathbf{dexp}_{\mathbf{x}}$ in (\ref{dexpSO3}), and rotation matrix $%
\mathbf{R}=\exp \tilde{\mathbf{x}}$ in (\ref{SO3exp1}) or (\ref{SO3exp2}),
where $\mathbf{x}\neq \mathbf{0}$ in (\ref{expSE32}) applies to finite pitch
motions (pure rotations and helical motions), and $\mathbf{x}=\mathbf{0}$ to
infinite pitch motions (pure translations).

The twist of body $i$ in body-fixed representation is determined by $\hat{%
\mathbf{V}}_{i}=\mathbf{C}_{i}^{-1}\dot{\mathbf{C}}_{i}\in se\left( 3\right) 
$, and in vector form $\mathbf{V}_{i}=\left( {^{i}}%
\bm{\omega}%
_{i},{^{i}}\mathbf{v}_{i}\right) $, where ${^{i}}\mathbf{v}_{i}=\mathbf{R}%
_{i}\dot{\mathbf{r}}_{i}$ is the linear velocity expressed in body-fixed
frame \cite{MUBOScrew1,Murray}. The time derivative of $\mathbf{X}$ and the
body-fixed twist are related by ${\mathbf{V}}=\mathbf{dexp}_{-\mathbf{X}}%
\dot{\mathbf{X}}$, respectively $\dot{\mathbf{X}}=\mathbf{dexp}_{-\mathbf{X}%
}^{-1}\mathbf{V}$. The right-trivialized differential and its inverse admit
closed form expressions \cite{RSPA2021}. The inverse is%
\begin{equation}
\mathbf{dexp}_{\mathbf{X}}^{-1}=\left( 
\begin{array}{cc}
\mathbf{dexp}_{\mathbf{x}}^{-1} & \ \ \mathbf{0} \\ 
\mathbf{B}%
\hspace{-0.4ex}%
\left( \mathbf{y}\right) & \mathbf{dexp}_{\mathbf{x}}^{-1}%
\end{array}%
\right)  \label{dexpInvSE3}
\end{equation}%
with $\mathbf{dexp}_{\mathbf{x}}^{-1}$ in (\ref{SO3dexpInv}) and 
\begin{equation}
\mathbf{B}%
\hspace{-0.4ex}%
\left( \mathbf{y}\right) =-\tfrac{1}{2}\tilde{\mathbf{y}}+\frac{1}{%
\left\Vert \mathbf{x}\right\Vert ^{2}}\left( 1-\gamma \right) \left( \tilde{%
\mathbf{x}}\tilde{\mathbf{y}}+\tilde{\mathbf{y}}\tilde{\mathbf{x}}\right) +%
\tfrac{\mathbf{x}^{T}\mathbf{y}}{\left\Vert \mathbf{x}\right\Vert ^{4}}%
\left( \tfrac{1}{\beta }+\gamma -2\right) \tilde{\mathbf{x}}^{2}.
\end{equation}

\paragraph{Cayley Map}

The Cayley map on $SE\left( 3\right) $ gives a rigid body displacement in
terms of the extended Rodrigues vector $\mathbf{X}=\left( \mathbf{c},\mathbf{%
d}\right) \in {\mathbb{R}}^{6}$ \cite%
{BauchauChoi2003,BauchauBook2010,RSPA2021}, where $\mathbf{c}\in {\mathbb{R}}%
^{3}$ is the Rodrigues vector for the rotation. It has the explicit form 
\cite{RSPA2021,Selig-IFToMM2007}%
\begin{equation}
\mathrm{cay}(\hat{\mathbf{X}})=\left( 
\begin{array}{cc}
\mathrm{cay}\left( \tilde{\mathbf{c}}\right)  & \left( \mathbf{I}+\mathrm{cay%
}\left( \tilde{\mathbf{c}}\right) \right) \mathbf{d} \\ 
\mathbf{0} & 1%
\end{array}%
\right)   \label{CaySE31}
\end{equation}%
with $\mathrm{cay}:so\left( 3\right) \mapsto SO\left( 3\right) $ in (\ref%
{CaySO31}). The inverse of its right-trivialized differential is%
\begin{equation}
\mathbf{dcay}_{\hat{\mathbf{X}}}^{-1}=\left( 
\begin{array}{cc}
\mathbf{dcay}_{\tilde{\mathbf{c}}}^{-1} & \mathbf{0} \\ 
\frac{1}{2}\left( \tilde{\mathbf{c}}-\mathbf{I}\right) \mathbf{d} & \frac{1}{%
2}\left( \mathbf{I}-\tilde{\mathbf{c}}\right) 
\end{array}%
\right)   \label{dCayInvSE3}
\end{equation}%
with $\mathbf{dcay}_{\tilde{\mathbf{c}}}^{-1}$ in (\ref{dCayInvSO32}) so
that $\mathbf{V}=\mathbf{dcay}_{\widehat{\mathbf{X}}}^{-1}\dot{\mathbf{X}}$.

\section{Interfacing Absolute Coordinate Models%
\label{secEmbed}%
}

Since in the past, absolute coordinate models have been developed for many
complex MBS, it is most natural aiming to make Lie group integrations
schemes applicable to such models. That is, when the classical formulations (%
\ref{EOMDyn})-(\ref{GeomConst}) or (\ref{EOMdyn2}),(\ref{EOMkin2}) of the
EOM are to be solved with Lie group methods, the absolute coordinates $%
\mathbf{q}\left( t\right) ,t\in \left[ t_{k},t_{k+1}\right] $ must be
computed from the absolute coordinates $\mathbf{q}\left( t_{k}\right) $ and
local coordinates $\mathbf{X}\left( t\right) $ during the integration step $%
t_{k+1}$. This is addressed in the remainder of this paper.

\subsection{The Local-Global Transition (LGT) Map}

\begin{figure}[h]
\vspace{-4ex} 
\centerline{
\includegraphics[height=6.3cm]{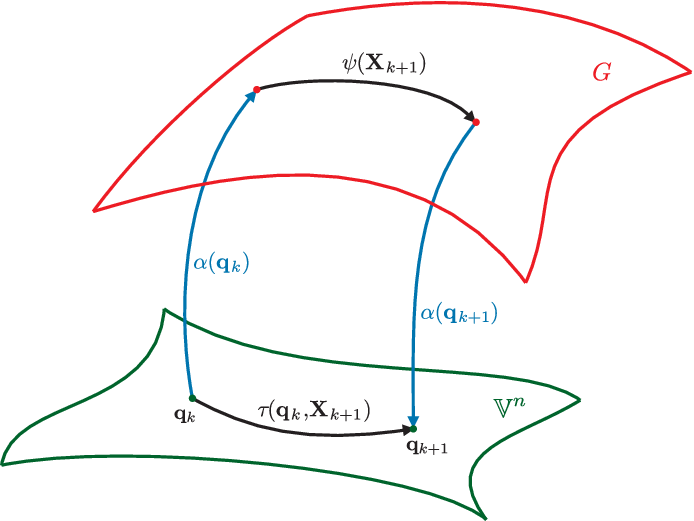}
}
\caption{Schematic view of the evaluation step using the LGT map.}
\label{figScheme}
\end{figure}

A Lie group integration scheme performs an update step (\ref{Step}) in terms
of local coordinates $\mathbf{X}\left( t\right) $. The dynamic EOM (\ref%
{LieEOMDyn4}) are evaluated with $C\left( t_{k}\right) $ computed from (\ref%
{Step}). They are related to the original EOM (\ref{EOMdyn2}) in absolute
coordinates in terms of the state $\left( \mathbf{q}\left( t\right) ,\mathbf{%
V}\left( t\right) \right) \in T{\mathbb{V}}^{n}$ via $C\left( t\right)
=\alpha \left( \mathbf{q}\left( t\right) \right) $. In order to apply a Lie
group integration scheme, the absolute coordinates $\mathbf{q}\left(
t\right) $ needed to evaluate the EOM for $t\in \left[ t_{k},t_{k+1}\right] $
within time step $k+1$ must be expressed in terms of the local coordinates $%
\mathbf{X}\left( t\right) $ along with the initial value $\mathbf{q}\left(
t_{k}\right) $. To this end, introduce the \emph{local-global transition
(LGT)} map%
\begin{eqnarray}
\tau &:&{\mathbb{V}}^{n}\times \mathfrak{g}\rightarrow {\mathbb{V}}^{n} \\
\left( \mathbf{q}\left( t_{k}\right) ,\mathbf{X}\left( t\right) \right)
&\mapsto &\mathbf{q}\left( t\right) =\tau \left( \mathbf{q}\left(
t_{k}\right) ,\mathbf{X}\left( t\right) \right)  \label{LGT}
\end{eqnarray}%
which determines the transition of the absolute coordinates from $\mathbf{q}%
\left( t_{k}\right) $ to the current $\mathbf{q}\left( t\right) $ according
to the incremental motion described by the local coordinates $\mathbf{X}%
\left( t\right) $. For general spatial MBS, $n=6N$. Figure \ref{figScheme}
shows a schematic view of the evaluation step using an LGT map. 

The LGT map can be used to two ends, as discussed next:

\begin{enumerate}
\item Integrating absolute coordinate models with Lie group methods,

\item Using the geometric model of rigid body motions, according to the Lie
group $G$, in absolute coordinate formulations.
\end{enumerate}

\subsection{Integrating Absolute Coordinate Models with Lie Group
Integration Schemes}

The LGT map (\ref{LGT}) allows formulating the EOM in terms of local
coordinates. Introducing it in (\ref{LieEOMDyn})-(\ref{LieVelConst}) yields
a model that only depends on $\mathbf{X}\left( t\right) $, and is
parameterized with $\mathbf{q}\left( t_{k}\right) $. This can be used as the
local parameter model within a Lie group integration for $t\in \left[
t_{k},t_{k+1}\right] $. For instance, when applying a Munthe-Kaas scheme,
the local model (\ref{LieEOMDyn3})-(\ref{LieEOMKin3}) in terms of $\mathbf{X}%
\left( t\right) ,\mathbf{V}\left( t\right) $ becomes%
\begin{eqnarray}
\dot{\mathbf{V}} &=&f_{k}\left( \mathbf{X}\left( t\right) ,\mathbf{V}\left(
t\right) ,t\right)  \label{locEOMdyn2} \\
\dot{\mathbf{X}} &=&\mathrm{d}\psi _{-\mathbf{X}\left( t\right) }^{-1}\left( 
{\mathbf{V}}\right) .  \label{locEOMkin2}
\end{eqnarray}%
with%
\begin{equation}
f_{k}\left( \mathbf{X}\left( t\right) ,\mathbf{V}\left( t\right) ,t\right)
:=f\left( \tau \left( \mathbf{q}\left( t_{k}\right) ,\mathbf{X}\left(
t\right) \right) ,\mathbf{V}\left( t\right) ,t\right) .  \label{fk}
\end{equation}%
This is an ODE system on the vector space ${\mathbb{R}}^{6n}\times {\mathbb{R%
}}^{6n}$ in terms of the state $\left( \mathbf{X}\left( t\right) ,\mathbf{V}%
\left( t\right) \right) $. It combines the original dynamics equations with
the Lie group modeling of kinematics, where the geometry of rigid body
motions is encoded in the LGT map used in (\ref{fk}) and the local
coordinate map $\psi $. Starting with initial conditions $\mathbf{X}\left(
t_{k}\right) =\mathbf{0},\mathbf{V}\left( t_{k}\right) $, and $\mathbf{q}%
\left( t_{k}\right) $, it can be solved for $\mathbf{X}\left( t\right) ,%
\mathbf{V}\left( t\right) ,t\in \left[ t_{k},t_{k+1}\right] $ with classical
vector space integration schemes. The solution for the absolute coordinates
at the time step $k+1$ is constructed as $\mathbf{q}\left( t_{k+1}\right)
=\tau \left( \mathbf{q}\left( t_{k}\right) ,\mathbf{X}\left( t_{k+1}\right)
\right) $.

\subsection{Geometrically Consistent Evaluation of Absolute Coordinate
Models integrated with Standard Vector Space Integration Schemes}

\begin{table*}[t] \centering%
\begin{tabular}{l|ll|}
\cline{2-3}
& \ \ \ \ \ \ \ \ \ \textbf{Absolute Coordinates for} & 
\hspace{-2ex}%
\textbf{Parameterization }$\alpha _{i}:{\mathbb{V}}^{\nu _{i}}\rightarrow
G_{i}$ \\ \cline{2-3}
& \textbf{1) Unit Quaternion + Position} & \multicolumn{1}{|l|}{\textbf{2)
Axis/Angle + Position}} \\ \cline{1-1}
\multicolumn{1}{|l|}{\textbf{Local coordinates}} & $\ \ \ \ \ \mathbf{q}%
_{i}=\left( \mathbf{Q}_{i},\mathbf{r}_{i}\right) \in Sp\left( 1\right)
\times {\mathbb{R}}^{3},\nu _{i}=7$ & \multicolumn{1}{|l|}{$\ \ \ \ \ 
\mathbf{q}_{i}=\left( 
\bm{\rho}%
_{i},\mathbf{r}_{i}\right) \in {\mathbb{R}}^{3}\times {\mathbb{R}}^{3},\nu
_{i}=6$} \\ \hline
\multicolumn{1}{|l|}{%
\begin{tabular}{l}
\textbf{a) Screw Coordinates} \\ 
$\ \ \ \ \ \mathbf{X}_{i}=\left( \mathbf{x}_{i},\mathbf{y}_{i}\right) \in {%
\mathbb{R}}^{6}\cong se\left( 3\right) $%
\end{tabular}%
} & \multicolumn{1}{|c}{$SE\left( 3\right) ,\exp $} & \multicolumn{1}{|c|}{$%
SE\left( 3\right) ,\exp $} \\ 
\multicolumn{1}{|l|}{%
\begin{tabular}{l}
\textbf{b) Axis/Angle + Position} \\ 
$\ \ \ \ \ \mathbf{X}_{i}=\left( \mathbf{x}_{i},\Delta \mathbf{r}_{i}\right)
\in {\mathbb{R}}^{3}\times {\mathbb{R}}^{3}$%
\end{tabular}%
} & \multicolumn{1}{|c}{$SO\left( 3\right) \times {\mathbb{R}}^{3},\exp $} & 
\multicolumn{1}{|c|}{$SO\left( 3\right) \times {\mathbb{R}}^{3},\exp $} \\ 
\multicolumn{1}{|l|}{%
\begin{tabular}{l}
\textbf{c) Rodrigues Parameters + Position} \\ 
$\ \ \ \ \ \mathbf{X}_{i}=\left( \mathbf{c}_{i},\Delta \mathbf{r}_{i}\right)
\in {\mathbb{R}}^{3}\times {\mathbb{R}}^{3}$%
\end{tabular}%
} & \multicolumn{1}{|c}{$SO\left( 3\right) \times {\mathbb{R}}^{3},\mathrm{%
cay}$} & \multicolumn{1}{|c|}{$SO\left( 3\right) \times {\mathbb{R}}^{3},%
\mathrm{cay}$} \\ 
\multicolumn{1}{|l|}{%
\begin{tabular}{l}
\textbf{d) Extended Rodrigues Parameters} \\ 
$\ \ \ \ \ \mathbf{X}_{i}=\left( \mathbf{c}_{i},\mathbf{d}_{i}\right) \in {%
\mathbb{R}}^{6}\cong se\left( 3\right) $%
\end{tabular}%
} & \multicolumn{1}{|c}{$SE\left( 3\right) ,\mathrm{cay}$} & 
\multicolumn{1}{|c|}{$SE\left( 3\right) ,\mathrm{cay}$} \\ \hline
\end{tabular}%
\caption{{Relevant combinations of $n_i=6$ local coordinates ${\bf X}_i$ with $\nu_i=6,7$ absolute coordinates ${\bf q}_i$ on $G_i$. 
The entries $G_i,\psi_i$ indicate the Lie groups and the local coordinate maps.}}%
\label{tabLGB}%
\end{table*}%
The LGT map can also be exploited for time integration of the absolute
coordinates model (\ref{EOMDyn})-(\ref{GeomConst}), respectively (\ref{EOM1}%
),(\ref{EOM2}). To this end, introduce the local model%
\begin{eqnarray}
\mathbf{M}\left( \mathbf{q}\right) \dot{\mathbf{V}}+\mathbf{A}^{T}%
\bm{\lambda}
&=&\mathbf{Q}\left( \mathbf{q},\mathbf{V},t\right)  \\
\dot{\mathbf{X}} &=&\mathrm{d}\psi _{-\mathbf{X}}^{-1}\left( {\mathbf{V}}%
\right)  \\
\mathbf{A}\left( \mathbf{q}\right) \mathbf{V} &=&\mathbf{0} \\
g\left( \mathbf{q}\right)  &=&\mathbf{0}
\end{eqnarray}%
in term of the (local) state $\mathbf{X}\left( t\right) ,\mathbf{V}\left(
t\right) $, with $\mathbf{q}\left( t\right) =\tau \left( \mathbf{q}\left(
t_{k}\right) ,\mathbf{X}\left( t_{k+1}\right) \right) $ for $t\in \left[
t_{k},t_{k+1}\right] $. It can be solved with any classical (multistep)
vector space integration method. This is advantageous when a decoupled
parameterization of spatial rigid body motions is used, i.e. when the
absolute rotations parameters do not affect translations, which corresponds
to $G=SO\left( 3\right) \times {\mathbb{R}}^{3}$. The latter is always the
case unless screw coordinates or dual quaternions are used as absolute
coordinates. Now in the above local model, the motion geometry, i.e. the Lie
group $G$, is encoded in the local coordinates and hence in the LGT. Thus, $%
G=SE\left( 3\right) $ can be used to model proper rigid body motions,
regardless of the absolute coordinates. The difference to integrating the
original model \ref{EOMDyn})-(\ref{GeomConst}) is that the integration
always starts with initial value $\mathbf{X}\left( t_{k}\right) =\mathbf{0}$%
. This comes with the condition that the integration scheme does not peruse
information from previous time steps, e.g. no solution extra-/interpolation
over multiple time steps, which would require computing $\mathbf{X}\left(
t\right) $ for $t<t_{k}$ from $\mathbf{q}\left( t_{k}\right) $.

\subsection{Relevant Local-Global Transition (LGT) Maps%
\label{secLGT}%
}

The formulation (\ref{locEOMdyn2}),(\ref{locEOMkin2}) admits using different
Lie groups to represent rigid body configurations, and using different
coordinate maps on the respective group. The LGT map is specific to the
absolute coordinates $\mathbf{q}_{i}$ used in the EOM (\ref{EOMDyn}), the
Lie group $G_{i}$ used within the integration time step, and the local
coordinates $\mathbf{X}_{i}$ (with corresponding $\psi $) on $G_{i}$. The
most widely used absolute coordinates are the following two:

\begin{enumerate}
\item[1)] $\mathbf{q}_{i}=\left( \mathbf{Q}_{i},\mathbf{r}_{i}\right) \in
Sp\left( 1\right) \times {\mathbb{R}}^{3}$, where $\mathbf{Q}_{i}=\left(
p_{0i},\mathbf{p}_{i}\right) $ is a unit quaternion describing the
orientation of the body-fixed frame at body $i$ relative to the IFR, and $%
\mathbf{r}_{i}$ is the position vector of the body-fixed frame resolved in
the IFR. This redundant parameterization is singularity-free. Notice that
the parameter space $Sp\left( 1\right) $ is itself a Lie group. In the EOM
formulation, unit quaternions are regarded as a 4-vector $\mathbf{Q}_{i}\in {%
\mathbb{R}}^{4}$ subjected to the unit norm constraint. Therefore, the
parameter space has dimension $\nu _{i}=7$.

\item[2)] $\mathbf{q}_{i}=\left( 
\bm{\rho}%
_{i},\mathbf{r}_{i}\right) \in {\mathbb{R}}^{6}$, where $%
\bm{\rho}%
_{i}$ scaled rotation vector (unit vector along the rotation axis multiplied
with the rotation angle), and $\mathbf{r}_{i}$ is the position vector, both
resolved in the IFR. This parameterization is not singularity-free. An
equivalent choice is the use of a three angle-descriptions (e.g. $%
\bm{\rho}%
_{i}$ could consist of three Euler angles).
\end{enumerate}

Local coordinates can be any canonical or non-canonical coordinates on the
chosen Lie group. The following four parameterizations are commonly used:

\begin{enumerate}
\item[a)] Screw coordinates $\mathbf{X}_{i}=\left( \mathbf{x}_{i},\mathbf{y}%
_{i}\right) \in {\mathbb{R}}^{6}\cong se\left( 3\right) $ as canonical
coordinates (of first kind); $G_{i}=SE\left( 3\right) $ and $\psi _{i}=\exp $%
.

\item[b)] Scaled rotation vector $\mathbf{x}_{i}\in {\mathbb{R}}^{3}\cong
so\left( 3\right) $ and position vector $\mathbf{r}_{i}\in {\mathbb{R}}^{3}$
as canonical coordinates $\mathbf{X}_{i}=\left( \mathbf{x}_{i},\mathbf{r}%
_{i}\right) \in {\mathbb{R}}^{3}\times {\mathbb{R}}^{3}$; $SO\left( 3\right)
\times {\mathbb{R}}^{3}$ and $\psi _{i}=\exp $.

\item[c)] Rodrigues-parameter $\mathbf{c}_{i}\in {\mathbb{R}}^{3}\cong
so\left( 3\right) $ and position vector $\mathbf{r}_{i}\in {\mathbb{R}}^{3}$
as non-canonical coordinates $\mathbf{X}_{i}=\left( \mathbf{c}_{i},\mathbf{r}%
_{i}\right) \in {\mathbb{R}}^{3}\times {\mathbb{R}}^{3}$; $SO\left( 3\right)
\times {\mathbb{R}}^{3}$ and $\psi _{i}=\mathrm{cay}$.

\item[d)] Extended Rodrigues-parameter $\mathbf{X}_{i}=\left( \mathbf{c}_{i},%
\mathbf{d}_{i}\right) \in {\mathbb{R}}^{6}\cong se\left( 3\right) $ as
non-canonical coordinates; $G_{i}=SE\left( 3\right) $ and $\psi _{i}=\mathrm{%
cay}$.
\end{enumerate}

The combinations of these absolute coordinates $\mathbf{q}_{i}$ and the
local coordinate maps $\psi $ (with coordinates $\mathbf{X}_{i}$) on the two
relevant choices for configuration space Lie groups $G_{i}$ are shown in
Table \ref{tabLGB}. The corresponding LGT maps are summarized in the
following.

\subsubsection*{\emph{1. Unit quaternion + position vector as absolute
coordinates}}

In most MBS dynamic codes using absolute coordinates, rigid body
orientations are described by unit quaternions, and positions are measured
in IFR, so that the absolute coordinates are $\mathbf{q}_{i}=\left( \mathbf{Q%
}_{i},\mathbf{r}_{i}\right) \in Sp\left( 1\right) \times {\mathbb{R}}^{3}$,
with $\nu _{i}=7$.

\subsubsection*{a) \emph{Screw coordinates as local coordinates}}

To capture rigid body motions, screw coordinates are used as local
coordinates $\mathbf{X}_{i}\in {\mathbb{R}}^{6}\cong se\left( 3\right) $,
which implies $G_{i}=SE\left( 3\right) $ and $\psi _{i}=\exp $. The index $i$
is omitted in the following. The LGT map $\tau :\left( Sp\left( 1\right)
\times {\mathbb{R}}^{3}\right) \times {\mathbb{R}}^{6}\rightarrow Sp\left(
1\right) \times {\mathbb{R}}^{3}$ for an individual body is separated as%
\begin{eqnarray}
\tau _{\mathrm{R}} &:&(Sp\left( 1\right) \times {\mathbb{R}}^{3})\times {%
\mathbb{R}}^{6}\rightarrow Sp\left( 1\right) ,\ \left( \mathbf{q},\mathbf{X}%
\right) \mapsto \mathbf{Q}^{\prime }=\tau _{\mathrm{R}}\left( \mathbf{q},%
\mathbf{X}\right) \ \ \ \ \ \ \ \   \label{LGT1a} \\
\tau _{\mathrm{T}} &:&(Sp\left( 1\right) \times {\mathbb{R}}^{3})\times {%
\mathbb{R}}^{6}\rightarrow {\mathbb{R}}^{3},\left( \mathbf{q},\mathbf{X}%
\right) \mapsto \mathbf{r}^{\prime }=\tau _{\mathrm{T}}\left( \mathbf{q},%
\mathbf{X}\right)  \notag
\end{eqnarray}%
where $\tau _{\mathrm{R}}$ determines the unit quaternion $\mathbf{Q}%
^{\prime }$ describing the absolute rotation due to the rigid body motion
within the time step parameterized by the screw coordinates $\mathbf{X}%
=\left( \mathbf{x},\mathbf{y}\right) $ starting from an initial rotation
described by the unit quaternion $\mathbf{Q}=\left( p_{0},\mathbf{p}\right) $%
, while $\tau _{\mathrm{R}}$ gives position vector $\mathbf{r}^{\prime }$
due to the motion when starting with position $\mathbf{r}$.

\paragraph{Mapping $\protect\tau _{\mathrm{R}}$:}

The scaled rotation vector $\mathbf{x}$ determines the incremental rotation,
which is represented by a unit quaternion using (\ref{Sp1exp}) as%
\begin{eqnarray}
\Delta \mathbf{Q} &\mathbf{=}&\left( \Delta p_{0},\Delta \mathbf{p}\right)
=\left( \cos \tfrac{\left\Vert \mathbf{x}\right\Vert }{2},\tfrac{1}{2}%
\mathrm{sinc}\tfrac{\left\Vert \mathbf{x}\right\Vert }{2}\mathbf{x}\right)
\label{DeltaQ} \\
&=&\left( \cos \tfrac{\varphi }{2},\mathbf{n}\sin \tfrac{\varphi }{2}\right)
\notag
\end{eqnarray}%
where $\mathbf{n}:=\mathbf{x}/\left\Vert \mathbf{x}\right\Vert $, and $%
\varphi :=\left\Vert \mathbf{x}\right\Vert $. The quaternion $\mathbf{Q}%
^{\prime }=\left( p_{0}^{\prime },\mathbf{p}^{\prime }\right) $ describing
the overall rotation is then found as%
\begin{equation}
\mathbf{Q}^{\prime }=\tau _{\mathrm{R}}\left( \mathbf{Q},\mathbf{x}\right)
\end{equation}%
with the mapping $\tau _{\mathrm{R}}$ given by the quaternion multiplication
(\ref{Qmult})%
\begin{align}
\tau _{\mathrm{R}}\left( \mathbf{Q},\mathbf{x}\right) :=& \;\mathbf{Q}\cdot
\Delta \mathbf{Q}\left( \mathbf{x}\right)  \notag \\
=& \;(p_{0}\Delta p_{0}\left( \mathbf{x}\right) -\mathbf{p}\cdot \Delta 
\mathbf{p}\left( \mathbf{x}\right) ,  \label{Qprime} \\
& \ \ \ \ \ \ \ \Delta p_{0}\left( \mathbf{x}\right) \mathbf{p}+p_{0}\Delta 
\mathbf{p}\left( \mathbf{x}\right) +\mathbf{p}\times \Delta \mathbf{p}\left( 
\mathbf{x}\right)  \notag \\
=& \;(p_{0}\cos \tfrac{\varphi }{2}-\tfrac{1}{2}\mathrm{sinc}\tfrac{%
\left\Vert \mathbf{x}\right\Vert }{2}\mathbf{p}^{T}\mathbf{x},  \notag \\
& \ \ \ \ \ \ \ \cos \tfrac{\varphi }{2}\mathbf{p}+\tfrac{1}{2}\mathrm{sinc}%
\tfrac{\left\Vert \mathbf{x}\right\Vert }{2}\left( p_{0}\mathbf{x}+\tilde{%
\mathbf{p}}^{2}\mathbf{x}\right) .  \notag
\end{align}

\paragraph{Mapping $\protect\tau _{\mathrm{T}}$:}

The incremental translation, relative to the initial displacement $\mathbf{r}
$, is deduced from (\ref{expSE31}) along with (\ref{dexpSO3}), in terms of
the screw coordinates $\mathbf{X}$ as%
\begin{eqnarray}
\Delta \mathbf{r}\left( \mathbf{X}\right) &=&\mathbf{dexp}_{\mathbf{x}}%
\mathbf{y}  \notag \\
&=&\left( \mathbf{I}+\tfrac{\beta }{2}\tilde{\mathbf{x}}+\left( 1-\alpha
\right) \tilde{\mathbf{n}}^{2}\right) \mathbf{y}  \label{Deltar}
\end{eqnarray}%
which is numerically robust also at $\left\Vert \mathbf{x}\right\Vert =0$.
The vector $\Delta \mathbf{r}$ is resolved in the body-fixed frame in its
initial orientation according to $\mathbf{Q}$ (since body-fixed twists, i.e.
left-trivialization is used). The absolute translation relative to IFR is
thus obtained by transforming the incremental displacement vector to IFR,
using the rotation matrix computed from the quaternion $\mathbf{Q}$ using (%
\ref{QuatR}), and hence $\mathbf{r}^{\prime }=\tau _{\mathrm{T}}\left( 
\mathbf{q},\mathbf{X}\right) $, with%
\begin{eqnarray}
\tau _{\mathrm{T}}\left( \mathbf{q},\mathbf{X}\right) {:=} &&\mathbf{r}+%
\mathbf{R}\left( \mathbf{Q}\right) \Delta \mathbf{r}\left( \mathbf{X}\right)
\notag \\
&=&\mathbf{r}+\Delta \mathbf{r}\left( \mathbf{X}\right) +2\left( p_{0}%
\widetilde{\mathbf{p}}+\widetilde{\mathbf{p}}^{2}\right) \Delta \mathbf{r}%
\left( \mathbf{X}\right) .  \label{rprime}
\end{eqnarray}%
It can also be computed with the quaternion transformation $\mathbf{r}%
^{\prime }=\mathbf{r}+\mathbf{Q}\cdot \Delta \mathbf{r}\cdot \mathbf{Q}%
^{\ast }$, where $\mathbf{Q}^{\ast }$ is the conjugate of $\mathbf{Q}$. In
summary the LGT map involves evaluating (\ref{Qprime}) and (\ref{Deltar})
followed by (\ref{rprime}). This determines the extra costs for using the
Lie group integration on $SE\left( 3\right) $ with screw coordinates as
local coordinates.

\subsubsection*{b) \emph{Relative rotation axis/angle and position vector as
local coordinates}}

With choice of local coordinates, rotations and translations are considered
decoupled, which implies the configuration space Lie group $G_{i}:=SO\left(
3\right) \times {\mathbb{R}}^{3}$. The coordinate map is the exponential in (%
\ref{expSO3R3}).

The LGT map determines $\mathbf{q}_{i}^{\prime }=\left( \mathbf{Q}%
_{i}^{\prime },\mathbf{r}_{i}^{\prime }\right) \in Sp\left( 1\right) \times {%
\mathbb{R}}^{3}$ from given absolute coordinates $\mathbf{q}_{i}=\left( 
\mathbf{Q}_{i},\mathbf{r}_{i}\right) \in Sp\left( 1\right) \times {\mathbb{R}%
}^{3}$ and scaled rotation vector $\mathbf{x}_{i}$ and incremental position
vector $\Delta \mathbf{r}_{i}$, which constitute the local coordinates $%
\mathbf{X}_{i}=\left( \mathbf{x}_{i},\Delta \mathbf{r}_{i}\right) \in {%
\mathbb{R}}^{3}\times {\mathbb{R}}^{3}$. Since the rotation and position
update are decoupled, the LGT map $\tau :\left( Sp\left( 1\right) \times {%
\mathbb{R}}^{3}\right) \times \left( {\mathbb{R}}^{3}\times {\mathbb{R}}%
^{3}\right) \rightarrow Sp\left( 1\right) \times {\mathbb{R}}^{3}$ on $G_{i}$
(i.e. for one body) splits as (omitting index $i$)%
\begin{align}
\tau _{\mathrm{R}}:(Sp\left( 1\right) \times {\mathbb{R}}^{3})\times \left( {%
\mathbb{R}}^{3}\times {\mathbb{R}}^{3}\right) & \rightarrow Sp\left( 1\right)
\label{LGT1b} \\
\left( \mathbf{Q},\mathbf{X}\right) & \mapsto \mathbf{Q}^{\prime }=\tau _{%
\mathrm{R}}\left( \mathbf{Q},\mathbf{x}\right)  \notag \\
\tau _{\mathrm{T}}:(Sp\left( 1\right) \times {\mathbb{R}}^{3})\times \left( {%
\mathbb{R}}^{3}\times {\mathbb{R}}^{3}\right) & \rightarrow {\mathbb{R}}^{3}
\notag \\
\left( \mathbf{q},\mathbf{X}\right) & \mapsto \mathbf{r}^{\prime }=\tau _{%
\mathrm{T}}\left( \mathbf{q},\Delta \mathbf{r}\right)  \notag
\end{align}

\paragraph{Mapping $\protect\tau _{\mathrm{R}}$:}

The LGT map $\tau _{\mathrm{R}}\left( \mathbf{Q},\mathbf{x}\right) $ for the
unit quaternion describing the absolute orientation remains as in (\ref%
{Qprime}).

\paragraph{Mapping $\protect\tau _{\mathrm{T}}$:}

Relation (\ref{rprime}) applies with the only difference that now $\Delta 
\mathbf{r}$ itself is part of the local coordinates. The absolute position
is thus $\mathbf{r}^{\prime }=\tau _{\mathrm{T}}\left( \mathbf{q},\Delta 
\mathbf{r}\right) $, with the LGT map%
\begin{eqnarray}
\tau _{\mathrm{T}}\left( \mathbf{q},\Delta \mathbf{r}\right) &{:=}&=\mathbf{r%
}+\mathbf{R}\left( \mathbf{Q}\right) \Delta \mathbf{r}  \notag \\
&=&\mathbf{r}+\Delta \mathbf{r}+2\left( p_{0}\widetilde{\mathbf{p}}+%
\widetilde{\mathbf{p}}^{2}\right) \Delta \mathbf{r}.  \label{rprime2}
\end{eqnarray}

For this parameterization, the LGT map involves evaluating (\ref{Qprime})
and (\ref{rprime2}). The computational expenditure is less than that when
using screw coordinates. If the motivation for using Lie group integration
is only to avoid singularities, then this could be preferable. Moreover,
this parameterization and corresponding update scheme have been proposed in 
\cite{TerzeMueller2016} for time integration of unit quaternions using
non-redundant coordinates, thus avoiding the renormalization of quaternions.

\subsubsection*{c) \emph{Rodrigues parameters and position vector as local
coordinates}}

Rodrigues parameters provide an algebraic description of rotations, i.e.
without trigonometric functions. They are not globally valid and cannot
describe full rotations, but are proper local coordinates. When combined
with unit quaternions as absolute coordinates, they yield an overall
algebraic parameterization, which may be computationally advantageous. The
configuration space Lie group corresponding to this parameterization is
again $G_{i}:=SO\left( 3\right) \times {\mathbb{R}}^{3}$, with $\psi _{i}=%
\mathrm{cay}$, and the LGT map splits as in (\ref{LGT1b}). The Rodrigues
parameters $\mathbf{c}_{i}$ and incremental position vector $\Delta \mathbf{r%
}_{i}$ constitute the local coordinates $\mathbf{X}_{i}=\left( \mathbf{c}%
_{i},\Delta \mathbf{r}_{i}\right) \in {\mathbb{R}}^{3}\times {\mathbb{R}}^{3}
$.

\paragraph{Mapping $\protect\tau _{\mathrm{R}}$:}

Unit quaternions are related to the Rodrigues parameters $\mathbf{c}\in {%
\mathbb{R}}^{3}$ by the gnomonic projection (\ref{QuatRodrigues}). The unit
quaternion for the incremental rotation described by Rodrigues parameters $%
\mathbf{c}\in {\mathbb{R}}^{3}$ is $\Delta \mathbf{Q}\left( \mathbf{c}%
\right) =\left( \Delta p_{0},\Delta \mathbf{p}\right) =(\sqrt{\sigma /2},%
\sqrt{1-\sigma /2}~\mathbf{c})$. The LGT map $\tau _{\mathrm{R}}\left( 
\mathbf{Q},\mathbf{c}\right) $ is thus (\ref{Qprime}) with $\Delta \mathbf{Q}%
\left( \mathbf{c}\right) $.

\paragraph{Mapping $\protect\tau _{\mathrm{T}}$:}

The LGT map $\tau _{\mathrm{T}}$ is again as in (\ref{rprime2}), which
determines the absolute position vector $\mathbf{r}^{\prime }=\tau _{\mathrm{%
T}}\left( \mathbf{q},\Delta \mathbf{r}\right) $.

\subsubsection*{d) \emph{Extended Rodrigues parameter as local coordinates}}

The Cayley map on $SE\left( 3\right) $, with extended Rodrigues parameters,
were used for geometrically exact modeling of flexible bodies in MBSs \cite%
{BauchauChoi2003,Bottasso1998,Borri2001b}. The geometric model is $%
G_{i}=SE\left( 3\right) $ and local coordinate map is $\psi _{i}=\mathrm{cay}
$ in (\ref{CaySE31}). The LGT map splits as in (\ref{LGT1a}), but now
extended Rodrigues parameters are used as local coordinates $\mathbf{X}%
_{i}=\left( \mathbf{c}_{i},\mathbf{d}_{i}\right) \in {\mathbb{R}}^{6}\cong
se\left( 3\right) $.

\paragraph{Mapping $\protect\tau _{\mathrm{R}}$:}

Since the $\mathbf{c}_{i}$ are Rodrigues parameters, the LGT map is the same
as in the above combination 1.c).

\paragraph{Mapping $\protect\tau _{\mathrm{T}}$:}

The incremental displacement is found with (\ref{CaySE31}) as $\Delta 
\mathbf{r}\left( \mathbf{X}\right) =\left( \mathbf{I}+\mathrm{cay}\left( 
\tilde{\mathbf{c}}\right) \right) \mathbf{d}$. The LGT map that gives the
absolute position vector is hence $\mathbf{r}^{\prime }=\tau _{\mathrm{T}%
}\left( \mathbf{q},\mathbf{X}\right) $ is $\tau _{\mathrm{T}}$ in (\ref%
{rprime}).

\subsubsection*{2. \emph{Rotation axis/angle and position vector as absolute
coordinates}}

Spatial rotations are described by the scaled rotation vector $%
\bm{\rho}%
\in {\mathbb{R}}^{3}$ and the absolute position by the vector $\mathbf{r}\in 
{\mathbb{R}}^{3}$ that constitute absolute coordinates $\mathbf{q}%
_{i}=\left( 
\bm{\rho}%
_{i},\mathbf{r}_{i}\right) \in {\mathbb{R}}^{3}\times {\mathbb{R}}^{3}$. The
scaled rotation vector is a representative of a 3-parametric description in
terms of canonical coordinates of first kind. Euler-angles, for instance,
are canonical coordinates of second kind on $SO\left( 3\right) $.

\subsubsection*{a) \emph{Screw coordinates as local coordinates}}

Rigid body motions are represented in $G_{i}=SE\left( 3\right) $, and screw
coordinates are used as local coordinates $\mathbf{X}_{i}=\left( \mathbf{x}%
_{i},\mathbf{y}_{i}\right) \in {\mathbb{R}}^{6}\cong se\left( 3\right) $,
with $\psi _{i}=\exp $ in (\ref{SO3exp1}). The LGT map $\tau :\left( {%
\mathbb{R}}^{3}\times {\mathbb{R}}^{3}\right) \times {\mathbb{R}}%
^{6}\rightarrow {\mathbb{R}}^{3}\times {\mathbb{R}}^{3}$ is split as
(omitting index $i$)%
\begin{eqnarray}
\tau _{\mathrm{R}} &:&\left( {\mathbb{R}}^{3}\times {\mathbb{R}}^{3}\right)
\times {\mathbb{R}}^{6}\rightarrow {\mathbb{R}}^{3},\ \left( \mathbf{q},%
\mathbf{X}\right) \mapsto 
\bm{\rho}%
^{\prime }=\tau _{\mathrm{R}}\left( 
\bm{\rho}%
,\mathbf{X}\right) \ \ \ \ \  \\
\tau _{\mathrm{T}} &:&\left( {\mathbb{R}}^{3}\times {\mathbb{R}}^{3}\right)
\times {\mathbb{R}}^{6}\rightarrow {\mathbb{R}}^{3},\left( \mathbf{q},%
\mathbf{X}\right) \mapsto \mathbf{r}^{\prime }=\tau _{\mathrm{T}}\left( 
\mathbf{q},\mathbf{X}\right) .  \notag
\end{eqnarray}

\paragraph{Mapping $\protect\tau _{\mathrm{R}}$:}

The LGT map $\tau _{\mathrm{R}}$ determines the scaled rotation vector for
the absolute rotation after incremental rotation according to $\mathbf{x}$.
That is, $%
\bm{\rho}%
^{\prime }$ is such that $\exp \tilde{%
\bm{\rho}%
}^{\prime }=\exp \tilde{%
\bm{\rho}%
}\exp \tilde{\mathbf{x}}$. The LGT map is thus given by the BCH formula (\ref%
{BCHSO3}) on $SO\left( 3\right) $, i.e. $\tau _{\mathrm{R}}\left( 
\bm{\rho}%
,\mathbf{x}\right) :=\mathrm{bch}\left( 
\bm{\rho}%
,\mathbf{x}\right) $.

\paragraph{Mapping $\protect\tau _{\mathrm{T}}$:}

The rotation matrix prior to the incremental rotation is computed as $%
\mathbf{R}\left( 
\bm{\rho}%
\right) =\exp \tilde{%
\bm{\rho}%
}$. The total absolute displacement vector after the incremental motion is
thus $\mathbf{r}^{\prime }=\tau _{\mathrm{T}}\left( \mathbf{q},\mathbf{X}%
\right) $, with LGT map%
\begin{equation}
\tau _{\mathrm{T}}\left( \mathbf{q},\mathbf{X}\right) :=\mathbf{r}+\mathbf{R}%
\left( 
\bm{\rho}%
\right) \Delta \mathbf{r}\left( \mathbf{X}\right)  \label{LGT2a}
\end{equation}%
with $\Delta \mathbf{r}\left( \mathbf{X}\right) $ in (\ref{Deltar}).

\subsubsection*{b) \emph{Rotation axis/angle and position vector as local
coordinates}}

Again $G_{i}:=SO\left( 3\right) \times {\mathbb{R}}^{3}$, with local
coordinates $\mathbf{X}_{i}=\left( \mathbf{x}_{i},\Delta \mathbf{r}%
_{i}\right) \in {\mathbb{R}}^{3}\times {\mathbb{R}}^{3}$ and $\psi =\exp $
in (\ref{expSO3R3}). The LGT map is%
\begin{align}
\tau _{\mathrm{R}}:\left( {\mathbb{R}}^{3}\times {\mathbb{R}}^{3}\right)
\times \left( {\mathbb{R}}^{3}\times {\mathbb{R}}^{3}\right) & \rightarrow {%
\mathbb{R}}^{3}  \label{LGT2b} \\
\left( \mathbf{q},\mathbf{X}\right) & \mapsto 
\bm{\rho}%
^{\prime }=\tau _{\mathrm{R}}\left( 
\bm{\rho}%
,\mathbf{x}\right)  \notag \\
\tau _{\mathrm{T}}:\left( {\mathbb{R}}^{3}\times {\mathbb{R}}^{3}\right)
\times \left( {\mathbb{R}}^{3}\times {\mathbb{R}}^{3}\right) & \rightarrow {%
\mathbb{R}}^{3}  \notag \\
\left( \mathbf{q},\mathbf{X}\right) & \mapsto \mathbf{r}^{\prime }=\tau _{%
\mathrm{T}}\left( \mathbf{q},\Delta \mathbf{r}\right)  \notag
\end{align}%
The rotation part is again $\tau _{\mathrm{R}}\left( 
\bm{\rho}%
,\mathbf{x}\right) :=\mathrm{bch}\left( 
\bm{\rho}%
,\mathbf{x}\right) $, and with $\mathbf{R}\left( 
\bm{\rho}%
\right) =\exp \tilde{%
\bm{\rho}%
}$, the translation part is%
\begin{equation}
\tau _{\mathrm{T}}\left( \mathbf{q},\mathbf{X}\right) :=\mathbf{r}+\mathbf{R}%
\left( 
\bm{\rho}%
\right) \Delta \mathbf{r}.  \label{LGT2bT}
\end{equation}

\subsubsection*{c) \emph{Rodrigues parameters and position vector as local
coordinates}}

Rodrigues parameters $\mathbf{c}_{i}$ and incremental position vector $%
\Delta \mathbf{r}_{i}$ are used as local coordinates $\mathbf{X}_{i}=\left( 
\mathbf{c}_{i},\Delta \mathbf{r}_{i}\right) \in {\mathbb{R}}^{3}\times {%
\mathbb{R}}^{3}$ with $\psi _{i}=\mathrm{cay}$ on $G_{i}:=SO\left( 3\right)
\times {\mathbb{R}}^{3}$. The LGT map $\tau _{\mathrm{T}}$ is the same as in
(\ref{LGT2b}). For rotation part, the LGT map is%
\begin{align}
\tau _{\mathrm{R}}:\left( {\mathbb{R}}^{3}\times {\mathbb{R}}^{3}\right)
\times \left( {\mathbb{R}}^{3}\times {\mathbb{R}}^{3}\right) & \rightarrow {%
\mathbb{R}}^{3}  \label{LGT2c} \\
\left( \mathbf{q},\mathbf{X}\right) & \mapsto 
\bm{\rho}%
^{\prime }=\tau _{\mathrm{R}}\left( 
\bm{\rho}%
,\mathbf{c}\right) .  \notag
\end{align}%
The absolute scaled rotation vector $%
\bm{\rho}%
^{\prime }$ is computed from $%
\bm{\rho}%
$ and $\mathbf{c}$ with the relation (\ref{LGT2cProof}).

\subsubsection*{d) \emph{Extended Rodrigues parameter as local coordinates}}

Extended Rodrigues parameter $\mathbf{X}=\left( \mathbf{c},\mathbf{d}\right) 
$ can be used as local coordinates, which represents a non-canonical
algebraic parameterization of $SE\left( 3\right) $. The map $\tau _{\mathrm{R%
}}$ is again (\ref{LGT2c}). The incremental translation is found from (\ref%
{CaySE31}) as $\Delta \mathbf{r}\left( \mathbf{X}\right) =\left( \mathbf{I}+%
\mathrm{cay}\left( \tilde{\mathbf{c}}\right) \right) \mathbf{d}$, and the
translation part $\tau _{\mathrm{T}}$ of the LGT map is given by (\ref%
{LGT2bT}), but with this $\Delta \mathbf{r}\left( \mathbf{X}\right) $.

\subsubsection*{Remarks on coordinate maps:}

Rodrigues parameters used as local coordinates in 1.c) and 2.c) cannot
describe full rotations, and further introduce an additional non-linearity
in the kinematics, which may interfere with the numerical integration if the
rotation within a time step becomes large. This can be mitigated by using
higher-order Cayley maps on $SO\left( 3\right) $ \cite%
{SchaubTsiotrasJunkins1995,TsiotrasJunkinsSchaub1997}. The second-order
Cayley map, for instance, leads to a description with Wiener-Milenkovic
parameter \cite{Milenkovic1982,Milenkovic2000}. The corresponding LGT map
can be derived straightforwardly. The same phenomenon occurs with the
extended Rodrigues parameters. This can also be addressed with higher-order
Cayley transformations on $SE\left( 3\right) $.

Unit coordinates are prevailing absolute coordinates in commercial MBS
codes. However, there are many implementations using canonical coordinates
(axis/angle, 3 angles), which can be integrated with combinations 2.a)-2.d).
Screw coordinates represent an alternative choice of absolute coordinates,
i.e. $\mathbf{q}_{i}=\left( \mathbf{x}_{i},\mathbf{y}_{i}\right) \in {%
\mathbb{R}}^{6}\cong se\left( 3\right) $ is the screw coordinate vector
represented in the IFR. This is, however, not beneficial from a
computational point of view, since this parameterization is not
singularity-free ($\mathbf{x}_{i}$ is again the scaled rotation axis).

\section{Conclusion%
\label{secConclusion}%
}

A formulation is presented that allows exploiting Lie group integration
schemes for singularity-free time integration of the EOM of MBS described in
terms of absolute coordinates. Various global coordinates can be used to
formulate the EOM. The method is applicable to general Lie group integration
methods. The crucial element is the LGT map according to the absolute
coordinates and the local coordinates used within the respective method. The
underlying configuration space Lie group can be chosen as deemed
appropriate. The LGT map can also be used to amend the absolute coordinate
formulation so that rigid body motions are represented by the respective Lie
group. This allows integrating absolute coordinate models with standard
vector space integration schemes.

Future work will address all possible combinations of absolute coordinates
in which the EOM are expressed with possible local coordinates on the Lie
groups, and the explicit derivation of the corresponding LGT mapping. The
proposed interfacing Lie group integration schemes to standard absolute
coordinate models inevitably involves additional computation of the LGT map.
The overall computational efficiency along with the accuracy will be
analyzed in detail, and experimentally validated.

\appendix%

\section{Derivation of the BCH Formula on $SO\left( 3\right) $}

The Baker-Campbell-Hausdorff (BCH) formula on $SO\left( 3\right) $ relates $%
\mathbf{x}_{1},\mathbf{x}_{2}\in {\mathbb{R}}^{3}$ to a third scaled
rotation vector $\mathbf{x}\in {\mathbb{R}}^{3}$, such that%
\begin{equation}
\exp \tilde{\mathbf{x}}=\exp \tilde{\mathbf{x}}_{1}\exp \tilde{\mathbf{x}}%
_{2}.  \label{BCH}
\end{equation}%
A closed form expression is derived for $SO\left( 3\right) $ making use of
the isomorphism $so\left( 3\right) \cong sp\left( 1\right) \cong {\mathbb{R}}%
^{3}$, and employing the multiplication of quaternions 
\begin{equation}
Q=Q_{1}\cdot Q_{2}=\left( q_{10}q_{20}-\mathbf{q}_{1}\cdot \mathbf{q}%
_{2},q_{10}\mathbf{q}_{2}+q_{20}\mathbf{q}_{1}+\mathbf{q}_{1}\times \mathbf{q%
}_{2}\right) .  \label{Qmult2}
\end{equation}%
These quaternions are given in terms of $\mathbf{x},\mathbf{x}_{1},\mathbf{x}%
_{2}\in {\mathbb{R}}^{3}$ by the exp map in (\ref{Sp1exp}). Inserting $%
q_{i0}=\cos \tfrac{\left\Vert \mathbf{x}_{i}\right\Vert }{2},\mathbf{q}_{i}=%
\tfrac{1}{2}\mathrm{sinc}\tfrac{\left\Vert \mathbf{x}_{i}\right\Vert }{2}%
\mathbf{x}_{i},i=1,2$ and $q_{0}=\cos \tfrac{\left\Vert \mathbf{x}%
\right\Vert }{2},\mathbf{q}=\tfrac{1}{2}\mathrm{sinc}\tfrac{\left\Vert 
\mathbf{x}\right\Vert }{2}\mathbf{x}$, and comparing elements in (\ref%
{Qmult2}) yields%
\begin{eqnarray*}
\cos \frac{\varphi }{2} &=&\cos \frac{\varphi _{1}}{2}\cos \frac{\varphi _{2}%
}{2}-\frac{1}{4}\mathrm{sinc}\varphi _{1}\mathrm{sinc}\varphi _{2}\mathbf{x}%
_{1}\cdot \mathbf{x}_{2} \\
\frac{1}{2}\mathbf{x}\mathrm{sinc}\frac{\varphi }{2} &=&\frac{1}{2}\mathrm{%
sinc}\frac{\varphi _{1}}{2}\cos \frac{\varphi _{2}}{2}\mathbf{x}_{1}+\frac{1%
}{2}\cos \frac{\varphi _{1}}{2}\mathrm{sinc}\frac{\varphi _{2}}{2}\mathbf{x}%
_{2} \\
&&+\frac{1}{4}\mathrm{sinc}\frac{\varphi _{1}}{2}\mathrm{sinc}\frac{\varphi
_{2}}{2}\mathbf{x}_{1}\times \mathbf{x}_{2}.
\end{eqnarray*}%
Solving these equations yields a solution $\mathbf{x}$ $=\mathrm{bch}\left( 
\mathbf{x}_{1},\mathbf{x}_{2}\right) $ of (\ref{BCH}) with the BCH formula%
\begin{equation}
\mathrm{bch\,}\left( \mathbf{x}_{1},\mathbf{x}_{2}\right) :=\alpha \mathbf{x}%
_{1}+\beta \mathbf{x}_{2}+\gamma \mathbf{x}_{1}\times \mathbf{x}_{2}
\label{BCHSO3}
\end{equation}%
where 
\begin{equation}
\alpha :=\frac{\mathrm{sinc}\frac{\varphi _{1}}{2}\cos \frac{\varphi _{2}}{2}%
}{\mathrm{sinc}\frac{\varphi }{2}},\beta :=\frac{\cos \frac{\varphi _{1}}{2}%
\mathrm{sinc}\frac{\varphi _{2}}{2}}{\mathrm{sinc}\frac{\varphi }{2}},\gamma
:=\frac{\mathrm{sinc}\frac{\varphi _{1}}{2}\mathrm{sinc}\frac{\varphi _{2}}{2%
}}{2\mathrm{sinc}\frac{\varphi }{2}}
\end{equation}%
and 
\begin{equation}
\varphi :=2\arccos \left( \cos \frac{\varphi _{1}}{2}\cos \frac{\varphi _{2}%
}{2}-\frac{1}{4}\mathrm{sinc}\frac{\varphi _{1}}{2}\mathrm{sinc}\frac{%
\varphi _{2}}{2}\mathbf{x}_{1}\cdot \mathbf{x}_{2}\right)
\end{equation}%
is the rotation angle of the compound rotation described by $\mathbf{R}=\exp 
\tilde{\mathbf{x}}$, respectively $\mathbf{Q}=\exp (0,\frac{1}{2}\mathbf{x})$%
. This expression was presented in \cite{ConduracheCiureanu2020}. A
different, but equivalent expressions were reported in \cite{Engo2001}.

\section{Derivation of the LGT map $\protect\tau _{\mathrm{R}}\left( 
\bm{\rho}%
,\mathbf{c}\right) $ in (\protect\ref{LGT2c})}

The derivation is similar to that of the BCH formula. Consider the
quaternion product $\mathbf{Q}\left( \mathbf{x}\right) =\mathbf{Q}_{1}%
\mathbf{\left( \mathbf{x}\right) Q}_{2}\left( \mathbf{c}\right) $, where $%
\mathbf{Q\left( \mathbf{x}\right) }=(\cos \tfrac{\varphi }{2},\tfrac{1}{2}%
\mathrm{sinc}\tfrac{\varphi }{2}\mathbf{x})$ and $\mathbf{Q}_{1}\mathbf{%
\left( \mathbf{x}\right) }=(\cos \tfrac{\varphi _{1}}{2},\tfrac{1}{2}\mathrm{%
sinc}\tfrac{\varphi _{1}}{2}\mathbf{x}_{1})$ are unit quaternions in terms
of axis and angle, and $\mathbf{Q}_{2}\left( \mathbf{c}\right) =\left( \sqrt{%
\sigma /2},\ \sqrt{1-\sigma /2}~\mathbf{c}\right) $ is a quaternion in terms
of Rodrigues parameter according to (\ref{QuatRodrigues}). Comparing the
terms in the quaternion product yields%
\begin{equation}
\mathbf{x}=\alpha \mathbf{x}_{1}+\beta \mathbf{c}+\gamma \mathbf{x}%
_{1}\times \mathbf{c}  \label{LGT2cProof}
\end{equation}%
with%
\begin{equation}
\alpha :=\frac{\sqrt{\frac{\sigma }{2}}\mathrm{sinc}\frac{\varphi _{1}}{2}}{%
\mathrm{sinc}\frac{\varphi }{2}},\beta :=\frac{2\sqrt{1-\frac{\sigma }{2}}%
\cos \frac{\varphi _{1}}{2}}{\mathrm{sinc}\frac{\varphi }{2}},\gamma :=\frac{%
\sqrt{1-\frac{\sigma }{2}}\mathrm{sinc}\frac{\varphi _{1}}{2}}{\mathrm{sinc}%
\frac{\varphi }{2}}
\end{equation}%
where%
\begin{equation}
\varphi :=2\arccos \left( \sqrt{\frac{\sigma }{2}}\cos \frac{\varphi _{1}}{2}%
-\frac{1}{2}\sqrt{1-\frac{\sigma }{2}}\mathrm{sinc}\frac{\varphi _{1}}{2}%
\mathbf{x}_{1}\cdot \mathbf{c}\right) .
\end{equation}

\section*{Acknowledgement}

This work was supported by the LCM-K2 Center within the framework of the
Austrian COMET-K2 program.\vspace{-3ex}

\bibliographystyle{IEEEtran}
\bibliography{LieGroupIntegration}

\begin{thebibliography}{10}
\providecommand{\url}[1]{#1}
\csname url@samestyle\endcsname
\providecommand{\newblock}{\relax}
\providecommand{\bibinfo}[2]{#2}
\providecommand{\BIBentrySTDinterwordspacing}{\spaceskip=0pt\relax}
\providecommand{\BIBentryALTinterwordstretchfactor}{4}
\providecommand{\BIBentryALTinterwordspacing}{\spaceskip=\fontdimen2\font plus
\BIBentryALTinterwordstretchfactor\fontdimen3\font minus
  \fontdimen4\font\relax}
\providecommand{\BIBforeignlanguage}[2]{{%
\expandafter\ifx\csname l@#1\endcsname\relax
\typeout{** WARNING: IEEEtran.bst: No hyphenation pattern has been}%
\typeout{** loaded for the language `#1'. Using the pattern for}%
\typeout{** the default language instead.}%
\else
\language=\csname l@#1\endcsname
\fi
#2}}
\providecommand{\BIBdecl}{\relax}
\BIBdecl

\bibitem{ShabanaBook}
A.~A. Shabana, \emph{Dynamics of Multibody Systems}, 4th~ed.\hskip 1em plus
  0.5em minus 0.4em\relax Cambridge University Press, 2013.

\bibitem{ArnoldBrulsCardona2015}
M.~Arnold, O.~Br{\"u}ls, and A.~Cardona, ``{Error analysis of
  generalized-$\alpha$ Lie group time integration methods for constrained
  mechanical systems},'' \emph{Numerische Mathematik}, vol. 129, no.~1, pp.
  149--179, 2015.

\bibitem{ArnoldHante2017}
M.~Arnold and S.~Hante, ``{Implementation details of a generalized-$\alpha$
  differential-algebraic equation Lie group method},'' \emph{Journal of
  Computational and Nonlinear Dynamics}, vol.~12, no.~2, 2017.

\bibitem{BrulsCardona2010}
O.~Br{\"u}ls and A.~Cardona, ``{On the use of Lie group time integrators in
  multibody dynamics},'' \emph{Journal of Computational and Nonlinear
  Dynamics}, vol.~5, no.~3, 2010.

\bibitem{BrulsCardonaArnold2012}
O.~Br{\"u}ls, A.~Cardona, and M.~Arnold, ``{Lie group generalized-alpha time
  integration of constrained flexible multibody systems},'' \emph{Mech. Mach.
  Theory}, vol.~34, pp. 121--137, 2012.

\bibitem{CelledoniOwren2003}
E.~Celledoni and B.~Owren, ``{Lie group methods for rigid body dynamics and
  time integration on manifolds},'' \emph{Computer Methods in Applied Mechanics
  and Engineering}, vol. 192, no. 3-4, pp. 421--438, 2003.

\bibitem{TerzeZlatarMueller2012}
Z.~Terze, D.~Zlatar, and A.~Mueller, ``{Lie-Group integration method for
  constrained multibody systems in stabilized DAE-index-1 form},''
  \emph{Multibody system dynamics}, 2012.

\bibitem{TerzeMueller2016}
Z.~Terze, A.~M{\"u}ller, and D.~Zlatar, ``Singularity-free time integration of
  rotational quaternions using non-redundant ordinary differential equations,''
  \emph{Multibody system dynamics}, vol.~38, no.~3, pp. 201--225, 2016.

\bibitem{TerzeZlatarMueller2015}
------, ``{Lie-group integration method for constrained multibody systems in
  state space},'' \emph{Multibody System Dynamics}, vol.~34, no.~3, pp.
  275--305, 2015.

\bibitem{MMTCSpace2014}
A.~M{\"u}ller and Z.~Terze, ``{The significance of the configuration space Lie
  group for the constraint satisfaction in numerical time integration of
  multibody systems},'' \emph{Mechanism and Machine Theory}, vol.~82, pp.
  173--202, 2014.

\bibitem{BIT2016}
A.~M{\"u}ller, ``A note on the motion representation and configuration update
  in time stepping schemes for the constrained rigid body,'' \emph{BIT
  Numerical Mathematics}, vol.~56, no.~3, pp. 995--1015, 2016.

\bibitem{BIT2016Erratum}
------, ``Erratum to: A note on the motion representation and configuration
  update in time stepping schemes for the constrained rigid body,'' \emph{BIT
  Numerical Mathematics}, vol.~56, no.~3, pp. 1017--1018, 2016.

\bibitem{SonnevilleCardonaBruls2014}
V.~Sonneville, A.~Cardona, and O.~Br{\"u}ls, ``{Geometrically exact beam finite
  element formulated on the special Euclidean group SE(3)},'' \emph{Computer
  Methods in Applied Mechanics and Engineering}, vol. 268, pp. 451--474, 2014.

\bibitem{BorriTrainelliBottasso2000}
M.~Borri, L.~Trainelli, and C.~L. Bottasso, ``On representations and
  parameterizations of motion,'' \emph{Multibody System Dynamics}, vol.~4,
  no.~2, pp. 129--193, 2000.

\bibitem{HolzingerGerstmayr2021}
S.~Holzinger and J.~Gerstmayr, ``Time integration of rigid bodies modelled with
  three rotation parameters,'' \emph{Multibody System Dynamics}, pp. 1--34,
  2021.

\bibitem{NikraveshBook1988}
P.~Nikravesh, \emph{Computer-aided analysis of mechanical systems}.\hskip 1em
  plus 0.5em minus 0.4em\relax Prentice Hall, 1988.

\bibitem{Blajer2001}
W.~Blajer, ``A geometrical interpretation and uniform matrix formulation of
  multibody system dynamics,'' \emph{ZAMM-Journal of Applied Mathematics and
  Mechanics/Zeitschrift f{\"u}r Angewandte Mathematik und Mechanik: Applied
  Mathematics and Mechanics}, vol.~81, no.~4, pp. 247--259, 2001.

\bibitem{Brauchli1991}
H.~Brauchli, ``Mass-orthogonal formulation of equations of motion for multibody
  systems,'' \emph{Zeitschrift f{\"u}r angewandte Mathematik und Physik ZAMP},
  vol.~42, no.~2, pp. 169--182, 1991.

\bibitem{TerzeNaudet2008}
Z.~Terze and J.~Naudet, ``Geometric properties of projective constraint
  violation stabilization method for generally constrained multibody systems on
  manifolds,'' \emph{Multibody System Dynamics}, vol.~20, no.~1, pp. 85--106,
  2008.

\bibitem{Barker1973}
L.~Barker, R.~Bowles, and L.~Williams, ``Development and application of a local
  linearization of quaternion rate equations in real-time flight simulation
  problems,'' \emph{NASA tech. note, TN D-7347}, 1973.

\bibitem{Nikravesh1985-1}
P.~E. Nikravesh, R.~Wehage, and O.~Kwon, ``Euler parameters in computational
  kinematics and dynamics. part 1,'' 1985.

\bibitem{Shabana2014}
A.~A. Shabana, ``Euler parameters kinetic singularity,'' \emph{Proceedings of
  the Institution of Mechanical Engineers, Part K: Journal of Multi-body
  Dynamics}, vol. 228, no.~3, pp. 307--313, 2014.

\bibitem{MollerGlocker2012}
M.~M{\"o}ller and C.~Glocker, ``Rigid body dynamics with a scalable body,
  quaternions and perfect constraints,'' \emph{Multibody system dynamics},
  vol.~27, no.~4, pp. 437--454, 2012.

\bibitem{HaghshenasJaryaniBowling2013}
M.~Haghshenas-Jaryani and A.~Bowling, ``{A new switching strategy for
  addressing Euler parameters in dynamic modeling and simulation of rigid
  multibody systems},'' \emph{{Multibody System Dynamics}}, vol.~30, pp.
  185--197, 2013.

\bibitem{ConduracheAASAIAA2017}
D.~Condurache, ``{Poisson-Darboux problems's extended in dual Lie algebra},''
  in \emph{AAS/AIAA Astrodynamics Specialist Conference, Stevenson, WA, USA},
  2017.

\bibitem{RSPA2021}
A.~M\"{u}ller, ``{Review of Exponential and Cayley Map on SE(3) as relevant for
  Lie Group Integration of the Generalized Poisson Equation and Flexible
  Multibody Systems},'' \emph{Royal Society Proceedings A}, 2021.

\bibitem{MUBOScrew1}
A.~M{\"{u}}ller, ``{Screw and Lie group theory in multibody dynamics --Motion
  representation and recursive kinematics of tree-topology systems},''
  \emph{Multibody System Dynamics}, vol.~43, no.~1, pp. 1--34, 2018.

\bibitem{Bottasso1998}
C.~L. Bottasso and M.~Borri, ``Integrating finite rotations,'' \emph{Computer
  Methods in Applied Mechanics and Engineering}, vol. 164, no. 3-4, pp.
  307--331, 1998.

\bibitem{Borri2001b}
C.~L. Bottasso, M.~Borri, and L.~Trainelli, ``{Integration of elastic multibody
  systems by invariant conserving/dissipating algorithms. II. Numerical schemes
  and applications},'' \emph{Computer Methods in Applied Mechanics and
  Engineering}, vol. 190, no. 29-30, pp. 3701--3733, 2001.

\bibitem{IserlesMuntheKaasNrsettZanna2000}
A.~Iserles, H.~Z. Munthe-Kaas, S.~P. N{\o}rsett, and A.~Zanna, ``Lie-group
  methods,'' \emph{Acta Numerica}, vol.~9, p. 215–365, 2000.

\bibitem{Owren2018}
E.~Celledoni, E.~{\c{C}}okaj, A.~Leone, D.~Murari, and B.~Owren, ``{Lie Group
  integrators for mechanical systems},'' \emph{arXiv preprint
  arXiv:2102.12778}, 2021.

\bibitem{MuntheKaas1998}
H.~Munthe-Kaas, ``{Runge-Kutta methods on Lie groups},'' \emph{BIT Numerical
  Mathematics}, vol.~38, no.~1, pp. 92--111, 1998.

\bibitem{MuntheKaas1999}
------, ``{High order Runge-Kutta methods on manifolds},'' \emph{Applied
  Numerical Mathematics}, vol.~29, no.~1, pp. 115--127, 1999.

\bibitem{WielochArnold2021}
V.~Wieloch and M.~Arnold, ``{BDF integrators for constrained mechanical systems
  on Lie groups},'' \emph{Journal of Computational and Applied Mathematics},
  vol. 387, p. 112517, 2021.

\bibitem{Newmark1959}
N.~Newmark, ``{A Method of Computation for Structural Dynamics},'' \emph{J.
  Engineering Mechanics Division ASCE}, no. 67-94, pp. 67--94, 1959.

\bibitem{Iserles1984}
A.~Iserles, ``Solving linear ordinary differential equations by exponentials of
  iterated commutators,'' \emph{Numerische Mathematik}, vol.~45, no.~2, pp.
  183--199, 1984.

\bibitem{FaltinsenMarthinsenMuntheKaas2001}
S.~Faltinsen, A.~Marthinsen, and H.~Z. Munthe-Kaas, ``Multistep methods
  integrating ordinary differential equations on manifolds,'' \emph{Applied
  numerical mathematics}, vol.~39, no. 3-4, pp. 349--365, 2001.

\bibitem{ChrouchGrossman1993}
P.~Crouch and R.~Grossman, ``Numerical integration of ordinary differential
  equations on manifolds,'' \emph{J. Nonlinear Sci.}, vol.~3, no.~1, pp. 1--33,
  1993.

\bibitem{OwrenMarthinsen2001}
B.~Owren and A.~Marthinsen, ``Integration methods based on canonical
  coordinates of the second kind,'' \emph{Numerische Mathematik}, vol.~87,
  no.~4, pp. 763--790, 2001.

\bibitem{Murray}
R.~Murray, Z.~Li, and S.~Sastry, \emph{A Mathematical Introduction to Robotic
  Manipulation}.\hskip 1em plus 0.5em minus 0.4em\relax CRC Press, 1994.

\bibitem{Selig-IFToMM2007}
J.~M. Selig, ``{Cayley maps for SE(3)},'' in \emph{12th International
  Federation for the Promotion of Mechanism and Machine Science World
  Congress}, 2007, p.~6.

\bibitem{CND2016}
A.~M\"{u}ller, ``Coordinate mappings for rigid body motions,'' \emph{ASME
  Journal of Computational and Nonlinear Dynamic}, vol. 12(2), 2016.

\bibitem{BauchauChoi2003}
O.~A. Bauchau and J.-Y. Choi, ``The vector parameterization of motion,''
  \emph{Nonlinear Dynamics}, vol.~33, no.~2, pp. 165--188, 2003.

\bibitem{BauchauBook2010}
O.~A. Bauchau, \emph{{Flexible Multibody Dynamics}}.\hskip 1em plus 0.5em minus
  0.4em\relax Springer Science \& Business Media, 2010.

\bibitem{SchaubTsiotrasJunkins1995}
H.~Schaub, P.~Tsiotras, and J.~L. Junkins, ``{Principal rotation
  representations of proper N$\times$ N orthogonal matrices},''
  \emph{International Journal of Engineering Science}, vol.~33, no.~15, pp.
  2277--2295, 1995.

\bibitem{TsiotrasJunkinsSchaub1997}
P.~Tsiotras, J.~L. Junkins, and H.~Schaub, ``{Higher-order Cayley transforms
  with applications to attitude representations},'' \emph{Journal of Guidance,
  Control, and Dynamics}, vol.~20, no.~3, pp. 528--534, 1997.

\bibitem{Milenkovic1982}
V.~Milenkovic, ``Coordinates suitable for angular motion synthesis in robots,''
  in \emph{Robots IV Conf. Proc, (SME)}.\hskip 1em plus 0.5em minus 0.4em\relax
  Society of Manufacturing Engineers, 1982, pp. 407--420.

\bibitem{Milenkovic2000}
V.~Milenkovic and P.~Milenkovic, ``{Unit Quaternion and CRV: Complementary
  Non-Singular Representations of Rigid-Body Orientation},'' in \emph{Advances
  in Robot Kinematics}.\hskip 1em plus 0.5em minus 0.4em\relax Springer, 2000,
  pp. 27--34.

\bibitem{ConduracheCiureanu2020}
D.~Condurache and I.-A. Ciureanu, ``{Baker--Campbell--Hausdorff--Dynkin Formula
  for the Lie Algebra of Rigid Body Displacements},'' \emph{Mathematics},
  vol.~8, no.~7, p. 1185, 2020.

\bibitem{Engo2001}
K.~Eng{\o}, ``{On the BCH-formula in so (3)},'' \emph{BIT Numerical
  Mathematics}, vol.~41, no.~3, pp. 629--632, 2001.

\end{thebibliography}

\end{document}